\documentclass[sigconf,nonacm]{acmart}
\usepackage[noEnd=true]{algpseudocodex}
\usepackage{algorithm,extdash,bbm,float}
\usepackage{amsmath} 
\usepackage{pgfplots,tikz,multirow,multicol,hhline,pifont,pgfplotstable}
\pgfplotsset{compat=1.18}
\usetikzlibrary{matrix}
\usepackage{subcaption}
\newcommand{\cmark}{\ding{51}}
\newcommand{\xmark}{\ding{55}}
\usepackage{enumitem}
\setlist[itemize]{align=parleft,left=0.1em..1.2em}
\usepackage{clipboard}
\usepackage{arydshln}
\usepackage{stfloats}

\title{QCore: Data-Efficient, On-Device Continual Calibration for Quantized Models---Extended Version}

\begin{document}

\author{\texorpdfstring{David Campos$^1$, Bin Yang$^2$, Tung Kieu$^1$, Miao Zhang$^3$, Chenjuan Guo$^2$, and Christian S. Jensen$^1$}{David Campos, Bin Yang, Tung Kieu, Miao Zhang, Chenjuan Guo, and Christian S. Jensen}}

\affiliation{%
  \institution{\texorpdfstring{$^1$Aalborg University, Denmark $^2$East China Normal University, China $^3$Harbin Institute of Technology, Shenzhen, China}{Aalborg University, Denmark East China Normal University, China Harbin Institute of Technology, Shenzhen, China}}
  \country {\texorpdfstring{$^1$\{dgcc,tungkvt,csj\}@cs.aau.dk $^2$\{byang,cjguo\}@dase.ecnu.edu.cn $^3$zhangmiao@hit.edu.cn}{dgcc,tungkvt,csj\}@cs.aau.dk \{byang,cjguo\}@dase.ecnu.edu.cn zhangmiao@hit.edu.cn}} \\
}

\renewcommand{\shortauthors}{David Campos et al.}
\settopmatter{printfolios=true}
\begin{abstract}

We are witnessing an increasing availability of streaming data that may contain valuable information on the underlying processes. It is thus attractive to be able to deploy machine learning models, e.g., for classification, on edge devices near sensors such that decisions can be made instantaneously, rather than first having to transmit incoming data to servers. To enable deployment on edge devices with limited storage and computational capabilities, the full-precision parameters in standard models can be quantized to use fewer bits. The resulting quantized models are then calibrated using back-propagation with the full training data to ensure accuracy. This one-time calibration works for deployments in static environments. However, model deployment in dynamic edge environments call for continual calibration to adaptively adjust quantized models to fit new incoming data, which may have different distributions with the original training data. 
The first difficulty in enabling continual calibration on the edge is that the full training data may be too large and thus cannot be assumed to be always available on edge devices. The second difficulty is that the use of back-propagation on the edge for repeated calibration is too expensive. We propose \texttt{QCore} to enable continual calibration on the edge. First, it compresses the full training data into a small subset to enable effective calibration of quantized models with different bit-widths. We also propose means of updating the subset when new streaming data arrives to reflect changes in the environment, while not forgetting earlier training data. Second, we propose a small bit-flipping network that works with the subset to update quantized model parameters, thus enabling efficient continual calibration without back-propagation. An experimental study, conducted with real-world data in a continual learning setting, offers insight into the properties of \texttt{QCore} and shows that it is capable of outperforming strong baseline methods.

\end{abstract}

\maketitle
\thispagestyle{plain}
\section{Introduction} \label{sec:introduction}

Due to developments such as the spread of the Internet of Things and the ongoing digitalization of societal and industrial processes, data streams that hold the potential to offer valuable insight into their underlying processes are becoming increasingly prevalent. To maximize value creation from such data, it is important to enable continual analytics and decision making on the edge devices that receive the data streams. For example, classification is important in applications such as health monitoring, autonomous driving, finances, and web services~\cite{RuizFLMB21}. The on-device deployment of such classification tasks can not only enhance the functionality of edge devices but can also reduce the dependency on external processing and yield improved efficiency and reduced classification latencies.

Increasingly sophisticated classification methods have emerged over the last decade~\cite{TanDBW22, abs-2304-13029}, with state-of-the-art methods often relying on large deep learning models~\cite{abs-2302-02515,HeZRS16} or even combinations of such models~\cite{MiddlehurstLFLB21}, thus posing high computational requirements~\cite{PanWZY0CGWTDZYZ23}.
These large models are typically unsuitable for edge deployment, where resources are limited. For example, in intelligent vehicle applications, in-vehicle controllers employ classification models to classify different driving statuses, but such controllers often have limited storage and support only low-bit integers, e.g., INT4 or INT8.

To enable the deployment of these models on edge devices with limited computational capabilities and storage, it is necessary to compress large classification models~\cite{DempsterSW21, 0002Z0KGJ23} through techniques such as model-parameter quantization (e.g., using 2, 4, or 8-bit representations)~\cite{0002Z0KGJ23}. However, this process relies on model calibration to maintain performance, which has two limitations that prevent its deployment on the edge.

\begin{figure*}[ht!]
\begin{subfigure}{0.45\linewidth}
        \includegraphics[width=0.95\linewidth]{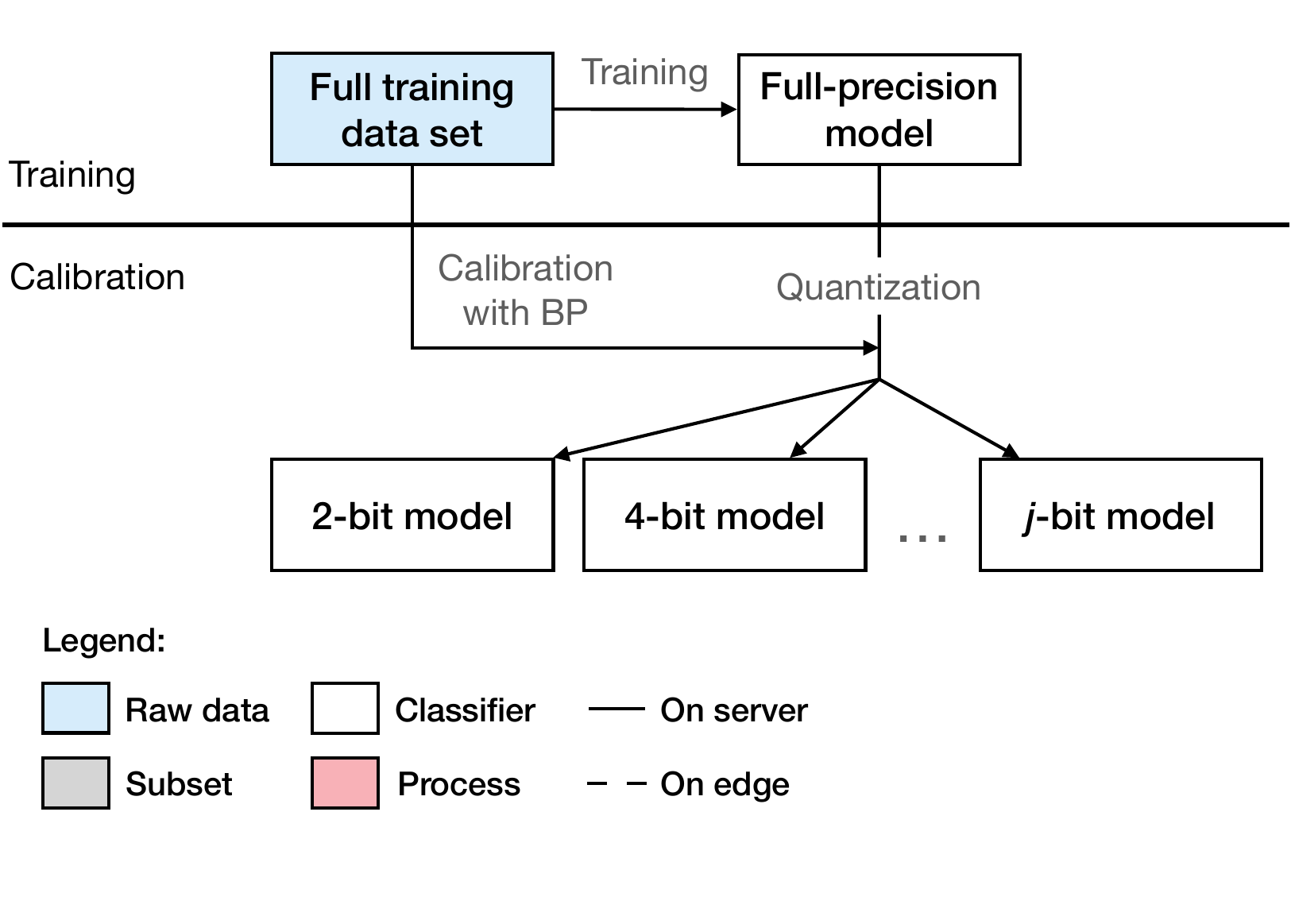}
        \vspace*{-0.5em}
    \caption{Traditional Quantization Paradigm. }
    \label{subfig:traditional}
\end{subfigure}
\begin{subfigure}{0.45\linewidth}
    \includegraphics[width=0.95\linewidth]{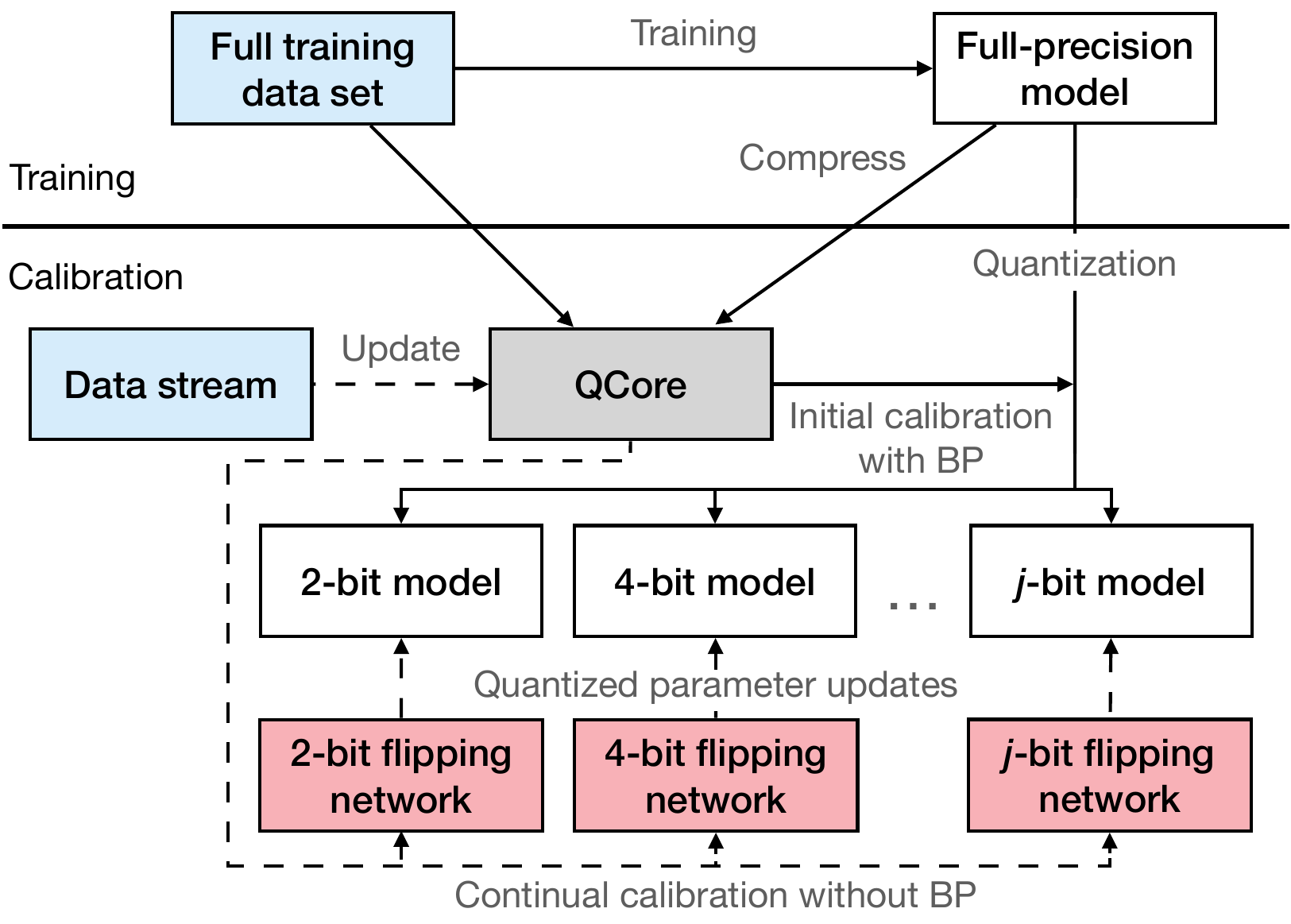}
    \vspace*{-0.5em}
    \caption{The \texttt{QCore} Paradigm.}
    \label{subfig:quantcore}
\end{subfigure}
    \caption{Paradigms for Quantized Classification. (a) In the traditional, one-time calibration paradigm, the full-training data set is required to perform calibration that uses back-propagation (BP). (b) In the proposed continual calibration paradigm, the full training data set is compressed into a small set, called \texttt{QCore}, that fits into edge devices with limited storage; and it is possible to update the \texttt{QCore} with incoming streaming data. Next, a bit-flipping network enables continual calibration without BP. }
    \label{fig:paradigms}
\end{figure*}

First, the calibration process is typically performed only once before deployment, using full training data and the full-precision model~\cite{NagelABLB20}, as shown in Figure~\ref{subfig:traditional}. 
In a streaming setting, this approach falls short because calibration needs to be executed continuously~\cite{haoicde24}. Further, edge devices may have insufficient storage to fit the full training data.
The continuous calibration is necessary because the classification occurs in dynamic environments rather than in static ones where one-time calibrations is sufficient. 

Specifically, in the targeted dynamic environments, distributions in incoming streaming data may vary considerably from what was seen in the original training data. For example, this may occur when vehicles are driven in varying climates or under varying driver behaviors and traffic conditions~\cite{PedersenYJ20,GuoYHJC20}. This necessitates re-calibration~\cite{0013FGD22,978-3-319-58347-1} that takes into account both past and new data.
For example, a vehicle with a driver-assistance system may adjust onboard sensor classifiers when changes in altitude and temperature occur, conditions that modify barometric pressure measurements~\cite{Mechtly1973}, or according to different driver behaviors~\cite{LiCJPGH22,Wu0ZG0J23}. 

Second, existing one-time calibration involves back-propagation (BP), which is computationally expensive and its accuracy relies on accurate computation of gradients in full-precision float values 
~\cite{GongLJLHLYY19}.  
However, once quantized models are deployed on the edge, full-precision float numbers may become unavailable as only lower-bit numbers, e.g., INT8, are available.
Moreover, performing back-propagation with low-bit parameters remains computationally costly, mainly due to extensive gradient computations~\cite{LiuM19}, and therefore impractical on edge devices. 

We proceed to summarize the limitations of the state-of-the-art that hinder the continual calibration of quantized models on the edge and then explain how we address these limitations.

\noindent
\textbf{Extensive Data Requirements: } 
Adjusting quantized models after deployment on edge devices requires substantial amounts of both initial training data and streaming data. This restricts such adjustment deployment on edge devices that, due to their limited storage, bandwidth, and computational capabilities, may be unable to store the data or may be inefficient. 
Therefore, we need means for reducing the data needed when calibrating quantized models on the edge, while still considering the original training data and streaming data to avoid catastrophic forgetting.
Further, these means must enable compliance with device memory size limitations. 

\noindent
\textbf{Lack of On-Edge Calibration: } 
Existing quantized model proposals do not support calibration once models are deployed. This limitation is particularly problematic in dynamic environments~\cite{LinCLCG020}. Calibrating a quantized model calls for minimizing a loss function by means of back-propagation, where full-precision computations are typically needed to estimate changes in the quantized model parameters. 
When a quantized model is deployed on the edge, back-propagation is less accurate due to the reduced precision of the quantized parameters. In addition, this process is costly due to the need for computing gradients for all parameters. This results in inefficiency and impracticality of conventional calibration on edge devices.
Instead, we need means of enabling continual calibration without access to full-precision parameters and without using back-propagation that relies on such parameters.

To eliminate the above limitations, we introduce \texttt{QCore}, a framework that supports the preparation, deployment,
and on-edge continual calibration of classification models on resource-limited edge devices.

\noindent
\textbf{Addressing Challenge 1: } 
To eliminate the data requirement limitation, we propose to compress the full training data set to a row-wise data subset, called \texttt{QCore}, designed to support quantized models calibration, thereby extending the traditional paradigm---see Figure~\ref{subfig:quantcore}. 
\texttt{QCore} supports one-time calibration when generating models with quantized parameters, and it is ready for subsequent on-edge continual calibrations. 
Compared to the traditional one-time model quantization, this process uses less data, enabling faster and more efficient deployment. Also, as it fits in edge devices, it can be updated as new data arrives, ensuring effective continual calibration that can balance past and new knowledge, as shown in the calibration step in Figure~\ref{subfig:quantcore}. This approach prevents the forgetting of past knowledge integrated it in a single data structure, whereas an additional buffer is typically required to achieve this goal in classic continual learning methods~\cite{RiemerCALRTT19}. 

Further, \texttt{QCore} is quantization-aware, meaning that it efficiently includes examples that support effective calibration of models with different levels of quantization that match the settings in which they are deployed, e.g., using different bits. This is important because there may be cases, at particular quantization levels, where models learn to classify data incorrectly, necessitating regular re-calibration within their quantization constraints.

\noindent
\textbf{Addressing Challenge 2: } 
To support continual calibration of quantized models after deployment, we propose an auxiliary so-called bit-flipping network. This network enables calibration of quantized parameter values in scenarios without access to full-precision values, while avoiding costly back-propagation. The bit-flipping network is designed to be compact to ensure deployment. Moreover, the bit-flipping network is quantized and exclusively conducts inference computations, minimizing the additional burdens on edge devices.
The bit-flipping network predicts whether a given quantized parameter value needs to be updated after processing incoming streaming data. The proposal represents a novel way of calibrating quantized models when back-propagation is too costly and full-precision parameter values are unavailable. The reliance on inferencing substantially reduces the computations necessary for updating a quantized model compared to using back-propagation. This makes the bit-flipping network highly attractive for on-edge calibration.

The bit-flipping network is integrated with \texttt{QCore} in order to update \texttt{QCore} as new data arrives. This ensures an effective continual learning process, allowing a model to maintain knowledge from the past while adapting to its environment. As a result, the model is capable of remaining competitive with all the data that has been processed, while address the challenges of working on the edge.

To the best of our knowledge, 
this is the first study of continual calibration without back-propagation of quantized models with diverse quantization levels on the edge.
We propose a method to obtain a compressed data set, called \texttt{QCore}, for calibrating models with different levels of quantization and a strategy for calibrating quantized model parameters. \texttt{QCore} is tailored for models with quantized parameters and work across the different stages of training a classification model, thus enabling for the creation of adjustable models suitable for edge devices. We also propose an innovative approach to enable calibration of quantized models when regular tools like back-propagation are infeasible due to limited resources. The paper makes the following contributions:

\begin{itemize}
\item It proposes \texttt{QCore}, a quantization-aware data set that compresses a full-training data set while identifying data examples that are important for the effective and efficient calibration of diverse bit-width quantized models.
\item It introduces an auxiliary lightweight network for efficient on-device learning of models with quantized parameters. This network eliminates costly back-propagation that requires full\Hyphdash precision parameters and gradient computations. Additionally, the network is quantized, making it suitable for use on edge devices.
\item It integrates \texttt{QCore} and the auxiliary network to continually adjust \texttt{QCore} to incoming data. This avoids the need for a buffer to prevent forgetting and instead uses a stable-sized data structure, which is essential for edge deployment.
\item It reports on extensive experiments that offer insight into the key design decisions and offer evidence of the applicability and effectiveness of \texttt{QCore} and the auxiliary network for on-device model deployment.
\end{itemize}

The paper is organized as follows. Section~\ref{sec:preliminaries} covers preliminaries, Section~\ref{sec:method} details the proposed method, and Section~\ref{sec:experiments} reports on experiments. Section~\ref{sec:related_work} reviews related work, and Section~\ref{sec:conclusions} concludes.

\section{Preliminaries} \label{sec:preliminaries}

This section presents concepts that are necessary to introduce the proposed framework.
\subsection{Classification Problem}

\subsubsection{Full Training Data Set}
A full training data set $\mathcal{D} = \{(x_i,y_i)\}^n_{i=1}$ is a collection of $n$ pairs $(x_i,y_i)$ defined over a $d$-dimensional feature space $\mathcal{X} \subset \mathbb{R}^d$ and a $k$-class label space $\mathcal{Y} \subset \mathbb{R}^k$. Each $x_i$ is an atomic entity to be classified, such as a time-series or an image represented as a $d$-dimensional vector, while the label $y_i$ indicates the specific class that the entity belongs to in the set of $k$ classes. 
For instance, in a human activity time-series data set, this set of labels includes conditions that represent different activity classes, such as walking, sitting, cycling, and running.

\subsubsection{Classification Task}
The classification task is to learn a function, or classifier, that takes an entity $x_i$ as input and returns its corresponding label $y_i$. During training, a classifier is learned using a full-training data set $\mathcal{D}$. The accuracy of the classifier is then evaluated on a testing data set $\mathcal{D}^\prime$ that is distinct from the full training data set $\mathcal{D}$.

When training a classifier with full-precision parameters $\Theta$, the objective is to learn the probability distribution $p^{\mathcal{D}}(\mathcal{Y}\mid\mathcal{X};\Theta)$ for all the pairs $(x_i,y_i)$ in full data training set $\mathcal{D}$, minimizing the cross-entropy loss between that probability and the ground truth $\mathcal{Y}$ captured by 
$
\arg \min_{\Theta} \mathcal{L}_{CE}(p^{\mathcal{D}}(\mathcal{Y}\mid\mathcal{X};\Theta), \mathcal{Y})
$.

\subsubsection{Streaming Batch}
A streaming batch is a data set $\mathcal{D}^t = \{(x_i^t,y_i^t)\}^m_{i=1}$ with $m \ll n$ that arrives at timestamp $t$, 
where each pair $(x_i^t,y_i^t) \in \mathcal{X} \times \mathcal{Y}$. The distribution of $\mathcal{X}$ may vary in $\mathcal{D}^t$ with respect to the training set $\mathcal{D}$, which would require to calibrate the classification model with respect to the known $\mathcal{Y}$ in $\mathcal{D}^t$.
For example, this may occur when different individuals perform activities that vary slightly due to factors like ages, health conditions, and changes in the environment, but still can be labeled by the original set of labels.

\subsubsection{Stream Classification Problem}

The classification problem, under streaming batches, becomes a continuous update of parameters $\Theta$ at time $t$, as captured by 
$
\arg \min_{\Theta^t} \mathcal{L}_{CE}(p(y_i^t \mid x_i^t;\Theta^{t-1}), y_i^t)
$.

\subsection{Quantization} \label{sec:quantized}

\begin{figure}[ht!]
    \centering
    \includegraphics[width=0.65\linewidth]{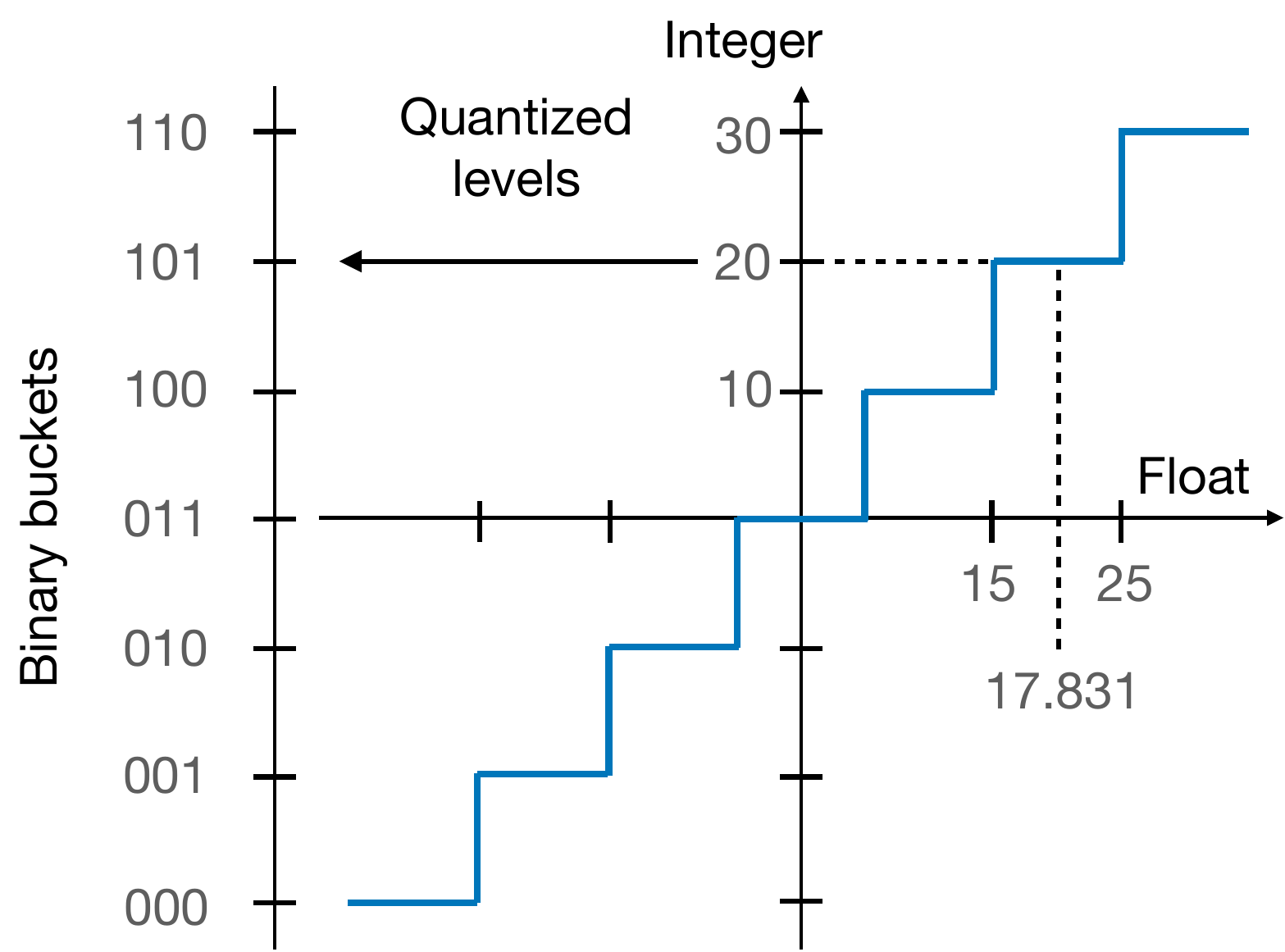}
    \caption{Quantization Mapping. }
    \label{fig:quantization}
\end{figure}

Quantization is a process that reduces the precision of model parameters, thus reducing model size. For instance, Figure~\ref{fig:quantization} shows how full-precision parameters, which are 32-bit floats, can be quantized into 3-bit parameters using uniform quantization~\cite{GongLJLHLYY19}.
In this example, the value 17.831 falls into the interval [15, 25) and so maps to 20, which is assigned to the 3-bit bucket 101.

\subsection{Calibration} \label{sec:calibration}

For a classification model with quantized parameters $\Theta_j$ at a quantization level $j$, the probability distribution needs to be relearned for the data set $\mathcal{D}$, which is known as calibration. The step is necessary because $\Theta_j$ has a loss of precision compared to $\Theta$ and the parameters that it represents are interdependent. Learning the parameters $\Theta_j$ involves computationally costly back-propagation, and the training data needed can require significant memory space, making it unsuitable for running on edge devices with limited computational resources, as summarized in Table~\ref{table:summary}. Also, the process usually relies on full-precision parameters to compute the gradients necessary for back-propagation. This is because the quantization functions discretize the values, making them not properly differentiable~\cite{LiuM19}, and leading to the zero-gradient problem~\cite{Lee0H21}.

\begin{table}[ht!]
\centering
\small
\caption{Calibration Optimization Comparison.}
\label{table:summary}
\begin{tabular}{|l|c|c|c|}
\hline
& \textit{Computation} & \textit{Memory Use} & \textit{Edge Ready} \\ \hline
Training Set + BP & High  & High & \xmark \\ \hline
\texttt{QCore} + BP & High & Low & \xmark \\ \hline
\texttt{QCore} + No BP & Low & Low & \cmark \\ \hline
\end{tabular}
\end{table}

To optimize the calibration process, we can reduce memory consumption and runs it faster by utilizing a representative subset, called \texttt{QCore}, $\mathcal{D}_c \subset \mathcal{D}$. However, it still requires the back-propagation step for learning the parameters $\Theta_j$, which makes it unsuitable for edge applications. Therefore, it is essential to remove the BP process in order to effectively prepare the calibration for execution on edge devices and to facilitate the its further development as a continual process. The goal is to find a function that effectively substitutes Equation~\ref{eq:bp} 
in order to compute $\Theta_j$ using the \texttt{QCore} $\mathcal{D}_c$.
\begin{gather}
    \arg \min_{\Theta_j} \mathit{BP}(\Theta_j \mid \mathcal{D}_c) = \arg \min_{\Theta_j} \mathcal{L}_{CE}(p^{\mathcal{D}_c}(\mathcal{Y}\mid\mathcal{X};\Theta_j), \mathcal{Y}) \label{eq:bp}
\end{gather}
\section{The QCore Method} \label{sec:method}

We present the problem setting and then proceed to present the components of our framework to support calibration for quantized models on the edge.

\subsection{Problem Setting}

We consider a setting where a large and already trained classification model needs to be deployed on edge devices with limited hardware capabilities. That is usually accomplished using a compression method, such as quantization, to reduce the model size, but
training data is still necessary for calibrating the resulting quantized models to maintain performance. 
The full-training data set may not be available on edge devices; and if it is, it may be too large to be used for calibration. However, if compressed, the full-training data set may still be used on resource-limited devices to calibrate the quantized models.

When quantized models are deployed on edge devices, they will likely operate in environments that differ from the ones they were trained on. As a result, it must be possible to adjust the data used to calibrate models in different deployments. Such adjustment enables better model calibration and facilitates the development of the calibration as a continual process. In Section~\ref{ssec:coreset}, we propose a data management strategy that aligns with the above scenario, selecting the most suitable examples for calibrating quantized models. The strategy involves saving a portion of the available data within a storage budget, called \texttt{QCore}. This is a subset $\mathcal{D}_c$ of $\mathcal{D}$ that serves as a proxy for the full-training data set $\mathcal{D}$. It compresses a large data set into a subset of rows instead of dimensions or columns~\cite{IlkhechiCGMFSC20,ElgamalYAMH15,YuAKPZCWI21,YangMQZ09,YuanSWYZY10}, allowing a given model or algorithm calibrated on \texttt{QCore} to produce results that approximate those produced when using the full-training data set for training.

Additionally, the traditional one-time calibration of quantized models is limited to the step when the quantized models are generated. This is because the calibration employs expensive back-propagation that requires 
the full-training data set to perform gradient computations. 
When quantized models are deployed on edge devices, further adjustments are therefore not considered. 
To address this limitation, we propose a scheme in Section~\ref{ssec:flipping} that enables the continual calibration of quantized models. This scheme does not need access to the original full-precision model, and it avoids expensive back-propagation. 

\begin{figure}[ht]
    \centering
    \includegraphics[width=0.85\linewidth]{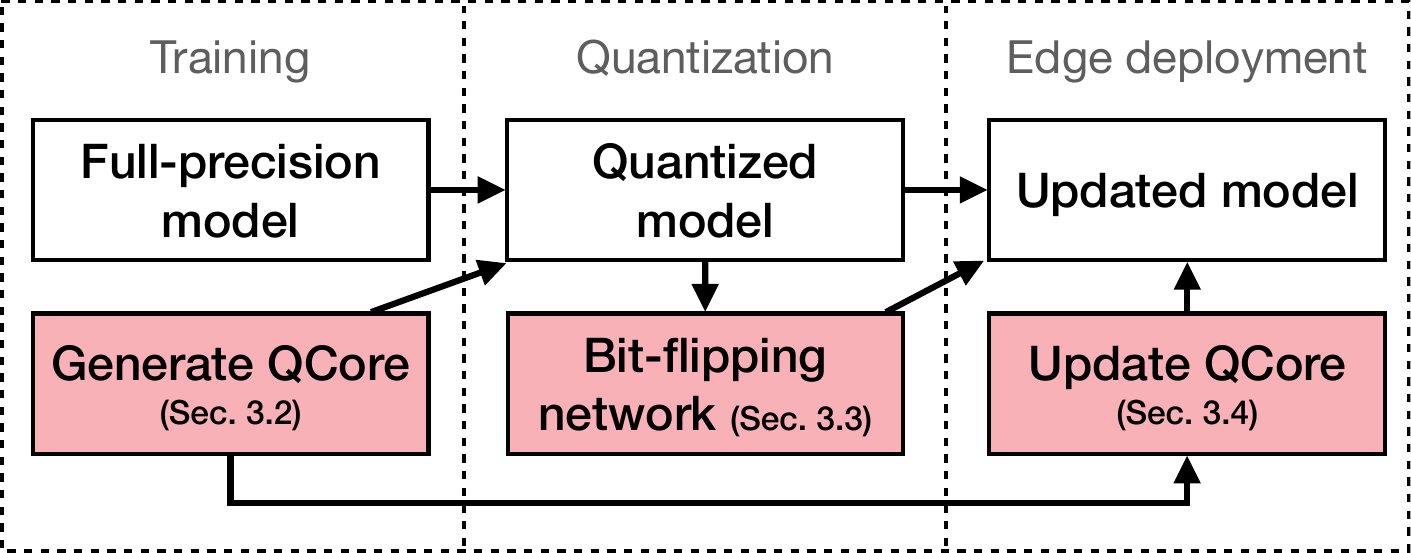}
    \caption{\texttt{QCore} Framework Overview. }
    \label{fig:overview}
\end{figure}

An overview of the proposal is shown in Figure~\ref{fig:overview}. First, given a full-precision classification model, \texttt{QCore} is designed to be able to calibrate quantized models with different quantization levels. This is essential as different edge devices may have different resource restrictions, thus requiring quantized models with different bit-widths. 
Next, the full-precision model is quantized based on its specific bit-width and a corresponding small bit-flipping network is trained for further continual calibration. Upon edge deployment, the quantized model is updated using the bit-flipping network, while \texttt{QCore} is updated with data obtained in the operating environment. 

Despite our focus being on classification, the framework is general and can be adapted to compress large models for other types of tasks, such as forecasting~\cite{xiaofeiTFB,pengiclr24,ZhaoGCHZY23,ChengCGZWYJ23,wupvldb,MileTS,razvanicde2021} and outlier detection models~\cite{KieuYGCZSJ22,KieuYGJZHZ22,davidpvldb}.

\subsection{Quantization-Aware Subset} \label{ssec:coreset}

\subsubsection{Subset Setting}

To handle the training data efficiently throughout the process of constructing quantized models, we propose a quantization-aware subset, called \texttt{QCore}. \texttt{QCore} serves the purpose of compressing the original training data considering the most suitable examples for calibrating quantized models, thereby reducing the data size. It has three important properties. First, it is small, making it easily implementable on edge devices. Second, it supports model calibration at different quantization levels, such as 2, 4, and 8 bits. Third, it can be adaptively updated after deployment in dynamic environments. 

Therefore, \texttt{QCore} is an essential component in the development and adjustment of quantized models for edge deployment. It plays a role in all stages of model quantization, as illustrated in the Figure~\ref{subfig:quantcore}. 
Initially, when a full-precision model is trained using the full-training data set, 
\texttt{QCore} is computed utilizing the full-precision model and the full-training data set, which are then available for the quantization step.
Then, according to different hardware restrictions, quantized models with different bits are quantized based on the full-precision model and are calibrated using the \texttt{QCore}, ready to be deployed on edge devices. As the models are utilized with data streams, each model can specialize its own \texttt{QCore} considering the changes introduced by each data stream, such as concept drifts. Each specific \texttt{QCore} can then be employed for calibrating its corresponding quantized model, allowing it to be tailored to the environment in which it is deployed.

A simple approach to building a subset is to randomly selecting a fraction of the full-training data set~\cite{PaparrizosLBHEE21}. However, this may lead to an unbalanced number of examples in terms of their utility for calibrating a quantized model. For instance, such a subset may contain an abundance of redundant and easy examples instead of more beneficial ones like boundary cases that are more useful for model calibration, 
as deep learning methods may process them differently~\cite{BaldockMN21,TonevaSCTBG19,KatharopoulosF18}.

In addition, when a full-precision model is compressed into models with varying quantization levels, such as 2, 4, or 8 bits, additional challenges arise. This is because data examples may have different significance when training models with different quantization levels. For example, certain data examples may be more challenging to be correctly classified in a 4-bit model than in an 8-bit model. 

Therefore, when compressing the full-training data set, it is necessary to assess the significance of each example for different quantized models. To do so, we consider empirical observations of the difficulty of every example when evaluated at different quantized levels while training the full-precision model. This is calculated using a metric that we called quantization misses.
Utilizing this metric, we identify relevant examples in the full-training data set that can effectively compress the data set and enable calibration of quantized models. The process is explained in further detail below.

\subsubsection{Quantization Misses} \label{sss:forgetting}

Consider a classification model with parameters $\Theta$ and the objective of learning the probability distribution $p(\mathcal{Y}\mid\mathcal{X};\Theta)$ for all pairs $(x_i,y_i)$ in the full-training data set $\mathcal{D}$. The predicted label for example $x_i$ at training step $s$ is denoted by $\hat{y}^s_i = \arg \max_k p(y_{ik}\mid x_i;\Theta^s)$, assuming $k$ classes.
The indicator function ${\mathit{TP}}_i^s$ returns a Boolean value that indicates whether example $x_i$ is correctly classified at step $s$:
%
\begin{equation}
    {\mathit{TP}}_i^s :=
    \begin{cases}
        1 ~&\text{ if }~ \hat{y}_i^s=y_i \\
        0 ~&\text{ otherwise}
    \end{cases}
    \label{eqn:indicator}
\end{equation}

A quantization miss for an example $x_i$ occurs when function ${\mathit{TP}}_i^s$ changes from 1 to 0 between consecutive training steps $s$ and $s+1$ when evaluating a given quantized model. This indicates that $x_i$ was classified correctly at step $s$, but misclassified at step $s+1$. By calculating the quantization misses for all examples during training of a specific quantized model, we can generate a probability mass function (PMF) for the full-training data set. This function represents the distribution of the examples within the training set in terms of quantization misses, providing an indicator of the difficulty of the training process for that specific quantized model. Extensively, when training a model and evaluating different quantized models, we obtain different distributions.

\begin{figure}[hb!]
\centering
    \includegraphics[width=0.85\linewidth]{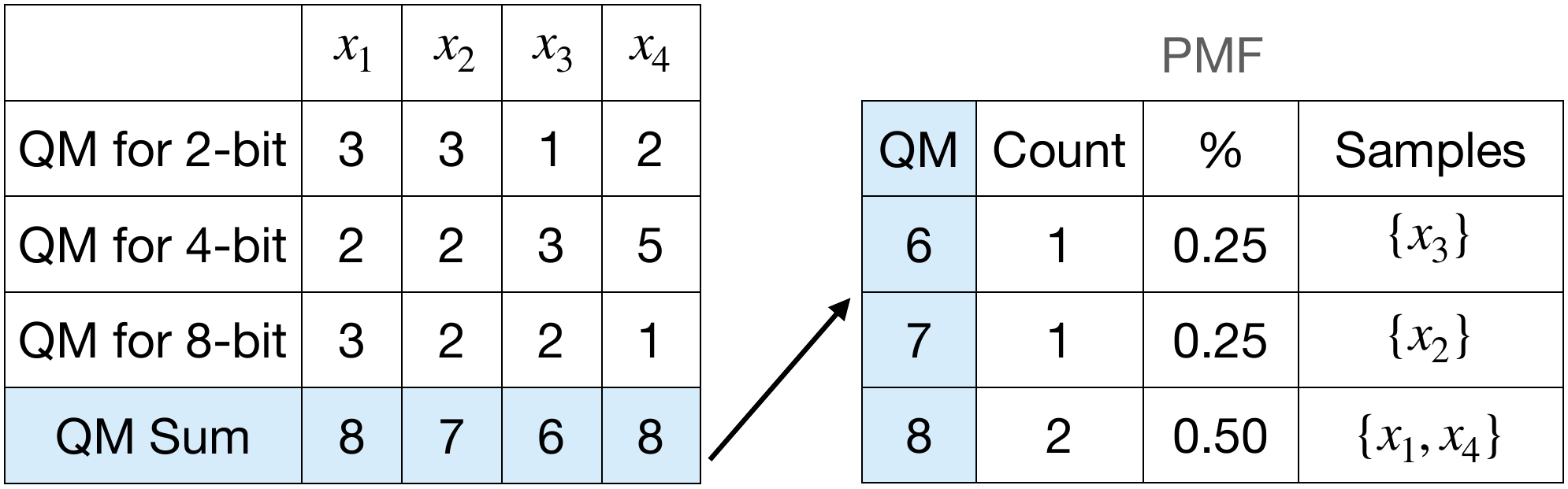}
    \caption{PMF Computation for Quantization Misses at Different Quantization Levels.}
    \label{fig:PMF_building}
\end{figure}

An example of how a PMF for quantization misses is generated is shown as follows. During training, quantization misses (denoted as QM in Figure~\ref{fig:PMF_building}) are counted for all data samples $x$ at each level of quantization, as shown to the left in Figure~\ref{fig:PMF_building} for four data samples and three quantization levels. 

\begin{figure}[ht!]
\footnotesize
\begin{subfigure}{0.5\linewidth}
    \pgfplotstableread{Figures/Data/PMF_A.txt}\coresetAdata
    \begin{tikzpicture}
        \begin{axis}[
            xlabel=Quantization misses,
            ylabel=Examples, 
            ymax=2500,
            ymode=log,
            ybar stacked,
            bar width=2.5mm,
            xtick scale label code/.code={},
            xtick = {1,2,3,4,5,6,7,8},
            xtick style={draw=none},
            enlarge x limits=0.15,
            width=1*\linewidth,
            height=0.52*\axisdefaultheight,
            legend pos= outer north east]
            \addplot[blue,fill=blue,mark=none, fill opacity=0.2] table[x expr=\coordindex+1, y=Coreset] {\coresetAdata};
            \addplot[gray,fill=gray,mark=none, fill opacity=0.2] table[x expr=\coordindex+1, y=Stacked] {\coresetAdata};
            \node at (3,750) {480};
            \node at (3,200) {$\downarrow$};
            \node at (3,70) {48};
        \end{axis}
    \end{tikzpicture}
    \vspace*{-0.5em}
    \caption{4-bit Quantized Model.}
    \label{subfig:forgetting_example_4}
\end{subfigure}
\hspace*{-1.5ex}
\begin{subfigure}{0.5\linewidth}
    \edef\windows{"970",}
    \pgfplotstableread{Figures/Data/PMF_C.txt}\coresetCdata
    \begin{tikzpicture}
        \begin{axis}[
            xlabel=Quantization misses,
            ylabel=Examples, 
            ymax=2500,
            ymode=log,
            ybar stacked,
            bar width=2.5mm, 
            xtick scale label code/.code={},
            xtick = {1,2,3,4,5,6,7},
            xtick style={draw=none},
            enlarge x limits=0.15,
            width=1*\linewidth,
            height=0.52*\axisdefaultheight,
            legend pos= outer north east]
            \addplot[purple,fill=purple,mark=none, fill opacity=0.2] table[x expr=\coordindex+1, y=Coreset] {\coresetCdata};
            \addplot[gray,fill=gray,mark=none, fill opacity=0.2] table[x expr=\coordindex+1, y=Stacked] {\coresetCdata};
            \node at (3,1350) {970};
            \node at (3,400) {$\downarrow$};
            \node at (3,140) {97};
        \end{axis}
    \end{tikzpicture}
    \vspace*{-0.5em}
    \caption{8-bit Quantized Model.}
    \label{subfig:forgetting_example_32}
\end{subfigure}
\caption{Distributions of Quantization Misses for Models with Different Precision. Subset Size $10\%$.} 
\label{fig:forgetting_example}
\end{figure}
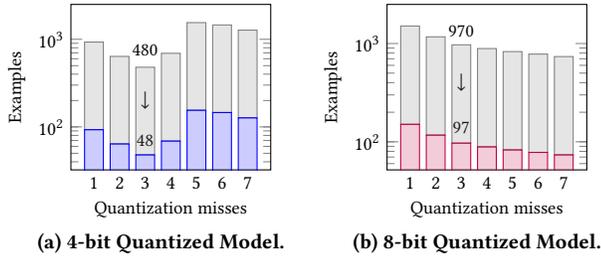

After training is completed, a PMF is generated for each quantization level by summing up all examples that share the same number of quantization misses, as shown to the right in Figure~\ref{fig:PMF_building}. 
Thus, the gray bars in Figure~\ref{fig:forgetting_example} represent the distributions of quantization misses obtained through 10,000 evaluations using a 4-bit and a 8-bit quantized models. The distributions differ noticeably between the two models, suggesting that certain examples pose more challenges for one model than the other, as the case with three quantization misses illustrates for 8-bit model.

\subsubsection{Generating QCore} \label{sssec:coreset_build}

Using the distribution generated by the quantization misses metric, we create a \texttt{QCore} by randomly selecting instances for each number of quantization misses. For example, in Figure~\ref{fig:forgetting_example} (note the logarithmic scale), we generate two \texttt{QCores} that are one-tenth the size of the training set. These \texttt{QCores} are shown as in the blue and red areas. They replicate the distribution of the full-training data set, but are much smaller, so the \texttt{QCore} $\mathcal{D}_c \sim \mathcal{D} \wedge \vert\mathcal{D}_c \vert
\ll \vert \mathcal{D} \vert$. In the case of the 4-bit model, the \texttt{QCore} includes 48 examples with three quantization misses, while the 8-bit model requires 97 examples at the same level.

We further extend the idea of computing the distribution of quantization misses to allow for flexibility in calibrating different quantized models. In doing so, we explore the possibility of combining the distributions of multiple quantized models into a single distribution. This enables us to showcase examples that, in general, pose greater difficulty when the model is quantized at multiple levels.
The process of generating \texttt{QCores} using quantization misses is integrated into full-precision model training, as shown in Algorithm~\ref{alg:generate_coreset}. 

\algblock{Input}{EndInput}
\algnotext{EndInput}
\algblock{Output}{EndOutput}
\algnotext{EndOutput}

\begin{algorithm}
\caption{Generate \texttt{QCore}.}
\label{alg:generate_coreset}
\begin{algorithmic}[1]

\Input : Full training data ($\mathcal{D}$), Full-precision network ($\mathit{FP}$), \texttt{QCore} size ($\mathit{Size}$) 
\EndInput
\Output : \texttt{QCore} ($\mathcal{D}_c$)
\EndOutput
\State $J \gets $ Quantization levels
\State $\mathit{QuantMisses}[\mathcal{D} \times J] \gets  \emptyset $ 
\Statex \tiny{\text{ }}
\For {$\mathit{s} \gets 1,\ldots,$ $\mathit{E}$} \label{line:startFP} \Comment{E epochs}
    \State $FP \gets $ Train $\mathit{FP}(\mathcal{D})$ 
        \For {$x_i \gets x_1,\ldots, x_N$}  \Comment{Every example in $\mathcal{D}$}  \label{line:batch}
            \For {$Q_j \gets $ Quantize $\mathit{FP}$ at quantization level $j \in J$}  
            \State $\hat{y}_{ij} \gets Q_j(x_i) $ \Comment{Inference with quantized model}
            \If {${\mathit{TP}}_{ij}^{s}$ changes from 1 to 0} \label{line:flips} \Comment{See Eq.~\ref{eqn:indicator}}
                \State $\mathit{QuantMisses}_j[x_i] \gets \mathit{QuantMisses}_j[x_i] + 1$ \label{line:register}
            \EndIf
        \EndFor
    \EndFor
\EndFor
\State \Comment{Count the $N_k^j$ examples with $k$ quantization misses at each $j$}
\State $\{(k, N_k^j) \gets \mathit{Distribute}(\mathit{QuantMisses}_j) : j \in J\}$ \label{line:distribution} \Comment{As Fig.~\ref{fig:forgetting_example}} \label{line:density}
\State $\{(k,N_k)\} \gets \sum_j N_k^j$ \Comment{Quantization Misses Distribution}
\State $\mathcal{D}_c \gets \mathit{Sample}(Size, \mathcal{D}, \{(k,N_k)\})$ \label{line:sampling}
\end{algorithmic}
\end{algorithm}

In each training step, the full-precision model is quantized temporary at different quantization levels to compute quantization misses. For example, the original model may be quantized at conventional power-of-two levels such as 2, 4, and 8, although other levels are also possible. This process is done online, meaning that the resulting quantized models are not adjusted any further and are replaced in subsequent executions. The temporary quantization step serves as a proxy between the full-precision model and the fully-trained quantized models. It enables estimation of how the quantized models will perform, enabling calculation of their quantization misses without undergoing training. Moreover, the challenging examples that are identified are likely to also pose difficulties after calibration, since they are already problematic for the proxy model. Importantly, this step also identifies the simple examples, helping to maintain a balanced distribution of training data between both categories.

The derived models are utilized to assess the examples for each batch and to predict their respective labels. Subsequently, each prediction is evaluated based on its outcome in the previous iteration to determine if it has transitioned from the correct label to an incorrect one. If an example has undergone a change in the evaluated model, a quantization miss for that specific example in the corresponding quantization level is noted.

After completing the training, 
a probability mass function (PMF) is generated for each quantization level by summing up all the examples based on their number of quantization misses.
This arrangement provides an outline of the distributions of training difficulty, as illustrated in Figure~\ref{fig:forgetting_example}. Specific \texttt{QCores} can be generated by sampling the training set using the distribution for each quantization level considered. However, a more general and flexible \texttt{QCore}, capable of supporting multiple quantization levels, considers the sum of the distributions, as shown in the last step of Algorithm~\ref{alg:generate_coreset}.

\noindent
\textit{Information loss:} 
We consider the $\epsilon$-approximation approach~\cite{FeldmanL11,reference/cg/2004,Badoiu08,JubranSNF21,RosmanVFFR14,Maalouf22}.

Given a full data set $\mathcal{D}$, this approach suggests that a coreset $\mathcal{D}_c \subset \mathcal{D}$ aims to minimize the difference of a cost function $\mathit{cost}(\cdot)$ that evaluates a given model $M$ on the full data set $\mathcal{D}$ vs. on the coreset $\mathcal{D}_c$. 
%
The resulting difference $\epsilon$, as shown in Equation~\ref{eq:cost},  
corresponds to the information loss when using the coreset $\mathcal{D}_c$ compared to using the full data set $\mathcal{D}$ on model $M$. 
Here, the cost function is, for example, a meta-function that is able to evaluate independently the performance of the model $M$ when the data changes~\cite{FeldmanL11,reference/cg/2004}. 

\begin{gather}
    \epsilon = \left|\frac{\sum_{x \in \mathcal{D}}cost(M,x)}{|\mathcal{D}|} - \frac{\sum_{x \in \mathcal{D}_c}cost(M,x)}{ |\mathcal{D}_c|}\right| 
    \label{eq:cost}
\end{gather}

In our setting, $M$ is a quantized model, and the cost function evaluates the number  
%
of quantization misses for a particular data point $x$ when calibrating $M$ w.r.t. a full-precision model. 
This choice is reasonable  
because 
the quantization misses thoroughly evaluate the quantized model $M$ during calibration, showing how the calibration of the quantized model evolves for each data point $x$. A similar concept, called ``forgetting events,'' is proposed to evaluate how the training of full precision models evolves~\cite{TonevaSCTBG19}. 
Thus, quantization misses based cost function can serve as a meta-function that shows a relationship between the quantized model $M$'s training status and the input data, which is  difficult to capture for deep learning models~\cite{Maalouf22}.

When using the full dataset $\mathcal{D}$ to calibrate model $M$, assume that $N_k$ data points exist who have $k$ quantization misses. 
Then we have 
\begin{equation}
\frac{\sum_{x \in \mathcal{D}} cost(M,x)}{|\mathcal{D}|}=\frac{\sum_{k=1}^K k \times N_k}{|\mathcal{D}|},
\label{eq:costD}
\end{equation}
where $K$ is 
the maximum level of quantization misses. 

Next, recall that we build our \texttt{QCore} $\mathcal{D}_c \subset \mathcal{D}$ by maintaining the same distribution of quantization misses but with a size $\lambda |\mathcal{D}|$, where $\lambda \in (0,1)$. Then, we have
\begin{equation}
\frac{\sum_{x \in \mathcal{D}_c} cost(M,x)}{|\mathcal{D}_c|}=
\frac{\sum_{k=1}^K k \times \lfloor \lambda N_k \rceil}{\lfloor \lambda |\mathcal{D}| \rceil}, 
\label{eq:costDC}
\end{equation}
where $\lfloor \cdot \rceil$ denotes the rounding operation. 

As Equations~\ref{eq:costD} and~\ref{eq:costDC} are both normalized, the normalized number of quantization misses should be the same. A possible difference is due to the non-proportional numbers of examples. 
First, $N_k$ must be an integer because it is the number of data points having $k$ quantization misses in $\mathcal{D}$. In contrast, $\lambda N_k$ may not be an integer, and when $\lfloor \lambda N_k \rceil$ is rounded, it incurs a rounding loss. The rounding loss is bounded to include at most one fewer or one more data point for each possible quantization miss. Thus, when summing up, the loss is at most the constant $K$, i.e., the maximum level of quantization misses. 
Thus, we have 
\begin{gather}
\left|\frac{\sum_{x \in \mathcal{D}} cost(M,x)}{|\mathcal{D}|}-\frac{\sum_{x \in \mathcal{D}_c} cost(M,x)}{|\mathcal{D}_c|}\right| \\
=\left|\frac{\sum_{k=1}^K k \times N_k}{|\mathcal{D}|}-
\frac{\sum_{k=1}^K k \times \lfloor \lambda N_k \rceil}{\lfloor \lambda |\mathcal{D}| \rceil}\right| \leq K 
\end{gather}
This shows that the information loss when using the \texttt{QCore} is bounded. 

In the following, we showcase the information loss analysis above using an example. Consider the example of Table~\ref{table:weights} with a full set $\mathcal{D}$ of size $|\mathcal{D}| = \sum N_k = 20$, a \texttt{QCore} $\mathcal{D}_c$ with 20\% of the size of the full data set (i.e., $\lambda=0.2$), and five possible quantization misses 1, 2, 3, 4, and 5. 

\begin{table}[ht!]
\addtolength{\tabcolsep}{-1pt}
    \small
    \centering
    \caption{Example of Information Loss Analysis, $\lambda=0.2$.}
    \label{table:weights}
    \begin{tabular}{ |*{6}{c|} } 
    \hline
& \multicolumn{2}{c|}{Full data set $\mathcal{D}$ } & \multicolumn{3}{c|}{\texttt{QCore} $\mathcal{D}_c$} \\ 
    \hline
$k$ & $N_k$ & $k \times N_k$ & $\lambda N_k$ & $\lfloor \lambda N_k\rceil $ & $k \times \lfloor \lambda N_k\rceil$ \\
\hline
1 & 2 & 2 & 0.4 & 0 & 0 \\ 
2 & 3 & 6 & 0.6 & 1 & 2 \\ 
3 & 9 & 27 & 1.8 & 2 & 6 \\ 
4 & 4 & 16 & 0.8 & 1 & 4 \\ 
5 & 2 & 10 & 0.4 & 0 & 0 \\ 
\hline
Total & 20 & 61 &  & 4 & 12 \\ 
\hline
    \end{tabular}
\end{table}

The normalized sum for quantization misses is: 

\begin{gather}
    \frac{\sum_{x \in \mathcal{D}} cost(M,x)}{|\mathcal{D}|} = \frac{\sum_{k=1}^{K=5} k\times N_k}{20} = \frac{61}{20} = 3.05 \label{eq:sum_D} \\
   \frac{\sum_{x \in \mathcal{D}_c} cost(M,x)}{|\mathcal{D}_c|} = \frac{\sum_{k=1}^{K=5} k \times \lfloor \lambda N_k \rceil}{4} = \frac{12}{4} = 3 \label{eq:sum_Dc} 
\end{gather}
Therefore, the information loss for the coreset $\mathcal{D}_c$ is $|3.05 -3| = 0.05$, which is bounded by $K=5$. 

\noindent
\textit{Complexity:} Algorithm~\ref{alg:generate_coreset}
executes $E$ training epochs over $N$ examples. In each epoch, back-propagation (BP) updates the model parameters $w$ with cost $BP_w$, which gives a training cost of $E \times N \times BP_w$.
Then, for every example in $N$, at each quantization level among all quantization levels $J$, the quantization misses counting can change at most $E$ times, since it is computed every epoch. Thus, the counting occurs at most $N\times J \times E$ times. 
To compute the distribution of quantization misses after training, we count the number the examples that have a specific number of quantization misses. This takes linear time w.r.t. $N$. 
This gives a cost of $E \times N \times BP_w + E \times N \times J + N$, where $J$ is often a small constant.
We obtain an asymptotic complexity of $\mathcal{O}(E \times N \times BP_w)$, which is the same as for regular back-propagation training.

\subsubsection{Calibrating Quantized Models} \label{sssec:coreset_use}

Upon model training and \texttt{QCore} generation, the model is ready for being compressed and deployed. 
Specifically, as \texttt{QCore} is tailored for quantized models, multiple versions with different bit-width levels can be generated and calibrated faster using the \texttt{QCore}, as the amount of data is reduced compared to the full training data. Any quantization strategy is applicable, consisting of applying a quantization function over the full-precision parameters and then using \texttt{QCore} to calibrate them, aiming to achieve similar performance to the original model.

Once the quantization process is complete, the calibrated models can be deployed on edge devices to perform inference tasks, while still having access to the \texttt{QCore}, which is small in size and thus can be stored on edge devices, for further continual calibration.

\subsection{Bit-flipping Network} \label{ssec:flipping}

\subsubsection{Overview}
Edge devices often encounter dynamic environments where the incoming data stream differs from the original training data. As a result, continual calibration at the edge becomes necessary in order to adapt the model. However, existing methods are impractical to execute at the edge because they depend on costly back-propagation. This method requires computing gradients for all the parameters and is inefficient due to its low performance caused by the loss of precision in the parameters.
To tackle this limitation, we introduce the bit-flipping network (BF), which is a small auxiliary quantized model with the same bit-width as the main quantized network. This network makes it possible to avoid computationally intensive tasks, such as back-propagation, effectively substituting it, i.e., Equation~\ref{eq:bp}, when trained in a given data set $\mathcal{D}_c$ to compute the parameters $\Theta_j$. 

To enable it to support calibration, the bit-flipping network is trained in parallel with the calibration for the quantized model, where the loss surface is consistent and stable, to calculate the expected changes to the parameters during that process. The bit-flipping network predicts one step ahead the potential changes to the parameters based on the current input to the classification model. The prediction is associated with only the three possible outcomes $\{-1,0,1\}$, and once the calibration is finished, the bit-flipping network can be employed during inferencing to forecast the parameter changes in the quantized model.

The bit-flipping network is a regression model that is trained alongside the main quantized model for a specific bit-width. This means that each model deployment has its own bit-flipping network. The bit-flipping network is only used for inference in edge devices. As a result, the computational cost is low, limiting the cost of when using the bit-flipping network on edge devices.

\subsubsection{Bit-flipping Network Training} \label{sssec:flipping_train}

During regular calibration, the model parameters are updated based on the input and how it affects the loss of the model. This update is performed using back-propagation. We have observed that there is a relationship between the input for each parameter and the actual change in the parameters after back-propagation. The input values for each parameter impact directly the magnitude of the change in that particular model parameter. 

\begin{figure}[b]
    \centering
    \includegraphics[width=0.9\linewidth]{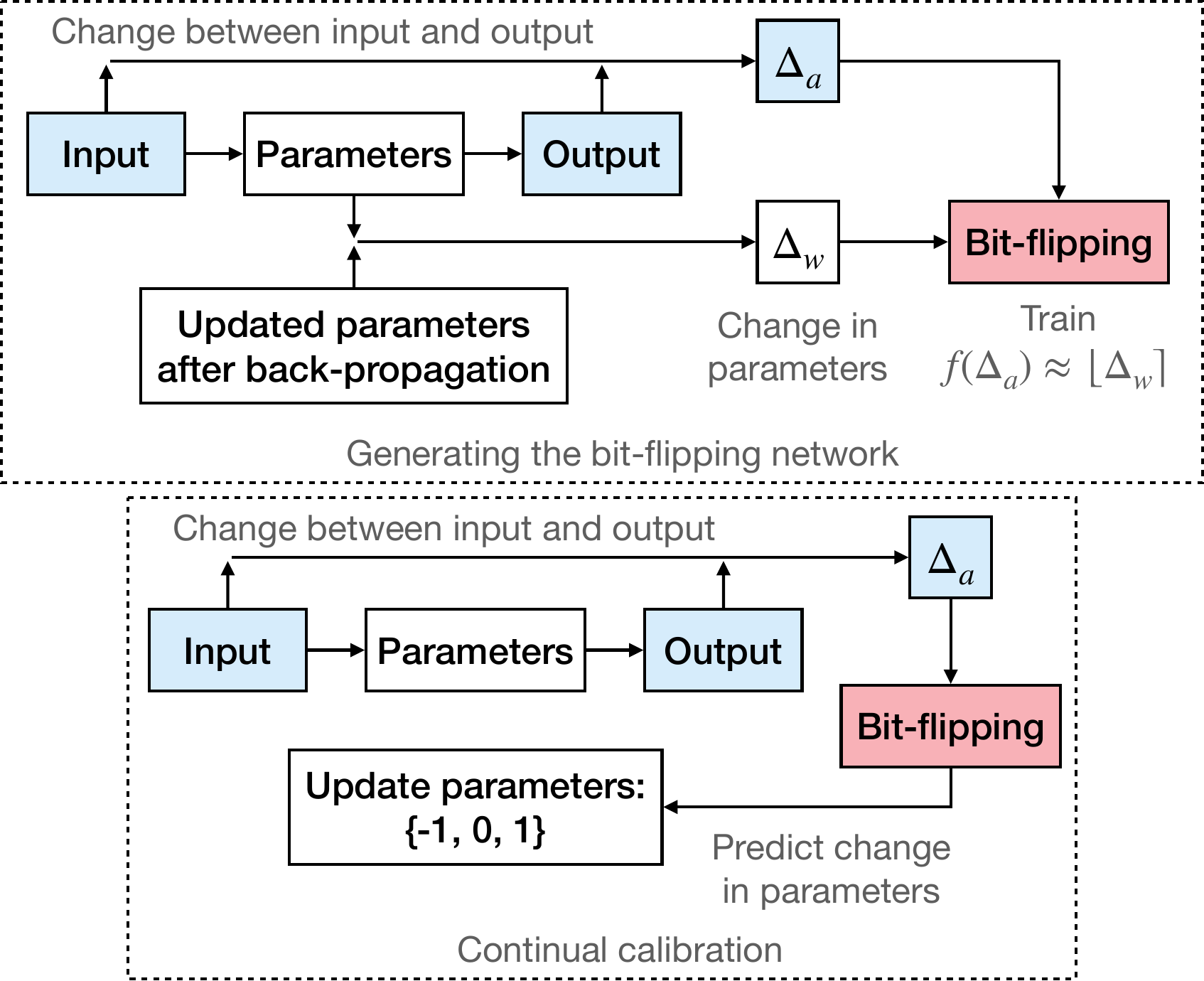}
    \caption{Bit-Flipping Training and On-Device Calibration.}
    \label{fig:bit-flipping}
\end{figure}

To study this relationship, as illustrated in the upper part of Figure~\ref{fig:bit-flipping}, we recorded the input and output of each parameter (shown as blue rectangles) during the first calibration of quantized models and calculated their difference as $\Delta_a$. Then, once the back-propagation step is computed, we calculate the change in the parameter as $\Delta_w$.

Using differences $\Delta_a$ and $\Delta_w$, we train the bit-flipping network to estimate the back-propagation results of the main model, conveyed as the change in parameter $\Delta_w$, considering the effect between the input and the parameter, represented by $\Delta_a$. The idea is to map the relationship between these two differences. For example, a higher value in input may lead the optimization process to reduce the value in the parameter, minimizing the effect of the high input value. Therefore, establishing this relationship allows us to identify how the model was trained originally when back-propagation was available and condense this knowledge into a small and efficient bit-flipping network.
Furthermore, as the training only considers the inputs for each parameter, a bit-flipping network trained on a given data set can be deployed to work with other data sets, even ones with different domains than the original model.

Then, because the main model is quantized, the potential changes in the parameters are only allowed to take on discrete values between -1 and 1, as strategies for changing bits have proven functional in calibrating quantized models~\cite{NagelABLB20,0003QLGZLY0G22}. This enables a simplified output from the bit-flipping network, which is limited to $\{-1,0,1\}$. As a result, the bit-flipping network can estimate whether a parameter is increasing, stays the same, or is decreasing without taking into account precise values like the gradients calculated during back-propagation.

Algorithm~\ref{alg:bit_flip_train} shows how the bit-flipping network is trained considering a quantized model $Q$ as a backbone. We keep the bit-flipping network architecture in a reduced size, consisting of a convolutional layer followed by a fully connected layer, quantized at the same bit-width level as the main model $Q$.
During the generation and calibration of quantized models using \texttt{QCore}, the difference between the input and output features of each parameter $w_i$, is computed and stored, as shown in line~\ref{line:activation}. Their computation depends on the incoming features from all the previous layers $w_k$, where $0\leq k<i$, denoted as $g\left(X \star \prod_{k=0}^{i-1} w_k^s\right)$, with an input $X$ and activation function $g$. 
Then, after back-propagation, the change in the parameters is also recorded.
The difference is maintained through discrete values that indicate whether the bits maintain their current values or change by one unit. We have observed that changes in the parameters are often within 1 bit, so we use a threshold to ensure that it stays within the range of -1 to 1.

\algblock{Input}{EndInput}
\algnotext{EndInput}
\algblock{Output}{EndOutput}
\algnotext{EndOutput}

\begin{algorithm}
\caption{Bit-flipping Training.}
\label{alg:bit_flip_train}
\begin{algorithmic}[1]

\Input : \texttt{QCore} ($\mathcal{D}_c$), Quantized Network ($\mathit{Q}$)
\EndInput
\Output : Bit-flipping Network ($\mathit{BF}$)
\EndOutput
\State $\Delta A [\mathit{Q.parameters} \times \mathit{E}] \gets  \emptyset, \quad \Delta{P} [\mathit{Q.parameters} \times \mathit{E}] \gets  \emptyset$
\State $X \gets \mathcal{D}_c$
\Statex \tiny{\text{ }}
\For {$\mathit{s} \gets 1,\ldots,$ $\mathit{E}$}
    \State $Q \gets $ Calibrate $Q(X)$ \Comment{See Section~\ref{sec:calibration}}
    \For {\text{each parameter} $w_i^s \text{ in } Q$}  
        \State $Act_{i} \gets g\left(X \star \prod_{k=0}^{i-1} w_k^s\right)$ \Comment{Input activation}
        \State $\Delta A[w_i^s,\mathit{s}] \gets (w_i^s \star Act_{i}) - Act_{i}$ \label{line:activation} 
        \State $w_i^{s+1} \gets w_i^s - \nabla Q(w_i^s)$ \Comment{BP update}
        \State $\Delta P [w_i^s,\mathit{s}] \gets \big\lfloor w_i^{s+1} - w_i^s \big\rceil \quad$   \Comment{The change is clipped and}
        \State \Comment{recorded as \{-1,0,1\}}
    \EndFor  
\EndFor
\vspace{0.5em}
\For {$\mathit{s} \gets 1,\ldots,$ $\mathit{E}$}
    \State $\mathit{BF} \gets $ Train $\mathit{BF}(\Delta A,\Delta P)$   \Comment{Train bit-flipping network} \label{line:bf_training}
\EndFor
\end{algorithmic}
\end{algorithm}

After recording all the changes in inputs and parameters during the calibration, we train the bit-flipping network using the input differences $\Delta_a$ as input and the resulting parameter differences $\Delta_w$ as their expected output, as shown in line~\ref{line:bf_training}.
This way, the bit-flipping network captures the effect on the input and subsequent parameter changes during the regular model calibration.

\noindent
\textit{Complexity:} 
As for Algorithm~\ref{alg:generate_coreset}, the calibration cost is $E \times N \times BP_w$. Then, its bit-flipping training takes $E \times BP_{bf}$, where $BP_{bf}$ has fewer parameters than $BP_{w}$. Thus, the asymptotic complexity remains the same as that of Algorithm~\ref{alg:generate_coreset}: $\mathcal{O}(E \times N \times BP_w)$.

\subsubsection{Bit-flipping Based Calibration} \label{sssec:flipping_update}

Once a classification model is deployed on an edge device, it can be calibrated using inference from the bit-flipping network. 
The calibration process is outlined in the lower part of Figure~\ref{fig:bit-flipping} and detailed in Algorithm~\ref{alg:bit_flip_update}, which runs together with Algorithm~\ref{alg:update_coreset}. 
First, the classification model is used to perform inference on \texttt{QCore} and incoming data, predicting their corresponding labels. 
Then, using \texttt{QCore} and the streaming data, the difference between the input and output features of each parameter in the main model is computed, as shown in line~\ref{line:activ_diff}. 
The result is used by the bit-flipping network to calculate the change in parameters, and update them in at most one unit. 

\algblock{Input}{EndInput}
\algnotext{EndInput}
\algblock{Output}{EndOutput}
\algnotext{EndOutput}

\begin{algorithm}
\caption{Bit-flipping Based Calibration.}
\label{alg:bit_flip_update}
\begin{algorithmic}[1]

\Input : \texttt{QCore} ($\mathcal{D}_c$), Stream batch ($\mathcal{D}_t$), Quantized model ($\mathit{Q}$), Bit-flipping network ($\mathit{BF}$)
\EndInput
\Output : Updated Quantized Network ($\mathit{Q}$)
\EndOutput
\State $X \gets \mathcal{D}_c \cup \mathcal{D}_t$
\Statex \tiny{\text{ }}
\For {$s \gets 1,\ldots,$ $\mathit{E}$}
    \For {\text{each parameter} $w_i^s \text{ in } Q$}
        \State $Act_{i}^s \gets g\left(X \star \prod_{k=0}^{i-1} w_k^s\right)$ \Comment{\texttt{QCore} updates in Alg.~\ref{alg:update_coreset}} \label{line:activ}
        \State $\Delta A_i^s \gets (w_i^s \star Act_{i}^s) - Act_{i}^s$  \label{line:activ_diff}
        \State $w_i^{s+1} \gets w_i^s + BF(\Delta A_i^s)$  \Comment{Update the parameters}
    \EndFor
\EndFor
\end{algorithmic}
\end{algorithm}

Because changing a parameter leads to modifications in the outputs of other parameters in the network, the process undergoes few iterations to ensure model stability. Subsequently, during calibration, different examples may exhibit quantization misses, which becomes the reason for updating \texttt{QCore} as outlined in Algorithm~\ref{alg:update_coreset} and explained in the following Section.

\noindent
\textit{Convergence:} 
We adapt established proofs for model optimization using back-propagation and gradient descent~\cite{Bubeck15,Lan2020} to consider quantized parameters. 

When using back-propagation to train a quantized model, the update rule for parameter $w_i$ at step $s$ is 
\begin{gather}
    w_i^{s}= Q(w_i^{s-1} - \eta \nabla f(w_i^{s-1})), \label{eq:bp_update1}
\end{gather}
where $Q(\cdot)$ is a quantization function that quantizes full-precision numbers, e.g., quantizing 32-bit floats to 4-bit integers, and 
$\eta$ is the learning rate. Thus, the update rule can be rewritten as 
\begin{gather}
    w_i^{s}=w_i^{s-1} - \eta \nabla f(w_i^{s-1}) + r_i^{s-1}, \label{eq:bp_update2} \\
    \text{where } r_i^{s-1} = Q(w_i^{s-1} - \eta \nabla f(w_i^{s-1})) - (w_i^{s-1} - \eta \nabla f(w_i^{s-1}))  \label{eq:bp_update3}
\end{gather}
Here, $r_i^{s-1}$ is the quantization error at step $s-1$, i.e., the difference between the quantized parameter and the full-precision parameter.

In our setting, the bit-flipping network $BF$ approximates the gradient and the quantization error as $BF(\cdot) \approx \eta \nabla f(w_i) - r_i$, so the update rule becomes
%
\begin{gather}
    w_i^{s}=w_i^{s-1} - BF(\cdot), \label{eq:bf_update1} \\
    \text{where } BF(\cdot) \approx \eta \nabla f(w_i^{s-1}) - r_i^{s-1} \label{eq:bf_update2} 
\end{gather}

The convergence analysis using the back-propagation based update rule, i.e., Equation~\ref{eq:bp_update2}, assumes that the variance of the gradients is bounded by a constant $G^2$: $\mathbb{E}||\nabla f(w_i)||^2 \leq G^2$. 
The condition is satisfied by using the $BF$ network based update rule, i.e., Equation~\ref{eq:bf_update1}, since $BF$ not only bounds the variance of $\nabla f(w_i)$ but the complete update component $\eta \nabla f(w_i) - r_i$. This is because 
the $BF$ 
only outputs values in $\{-1,0,1\}$, so its variance is bounded by one, as $\mathbb{E}||BF(\cdot)||^2 = 1$.

Then, following the standard convergence analysis for neural networks~\cite{Bubeck15,Lan2020} and given $\mathbb{E}||BF(\cdot)||^2=1$, we obtain the convergence rate over $m$ iterations as
\begin{gather}
    \mathbb{E}[f(\overline{w}_i) - f(w_i^*)] \leq \frac{1}{2m} \mathbb{E}||w_i^{0} - w_i^{*}||^2 + \frac{1}{2}, \label{eq:norm}
\end{gather}
where $\overline{w}_i = \frac{1}{m} \sum^m w_i^s$ is the average parameter, $w_i^{0}$ is the non-calibrated parameter, and $w_i^{*}$ is the optimal parameter.

\noindent
\textit{Complexity:} Algorithm~\ref{alg:bit_flip_update} has cost $E \times |Q| \times BF$, where $|Q|$ is the number of parameters of the quantized model and $BF$ is the inference cost of the bit-flipping network. As the size of the streaming batch and \texttt{QCore} is at most $N$, and as the number of parameters of bit-flipping network is $BF_w$, $BF$ is $N \times BF_w$. Therefore, its asymptotic complexity is $\mathcal{O}(E \times |Q| \times N \times BF_w)$.

\subsection{QCore Update} \label{ssec:coreset_update}

\begin{figure}[ht]
    \centering
    \includegraphics[width=0.9\linewidth]{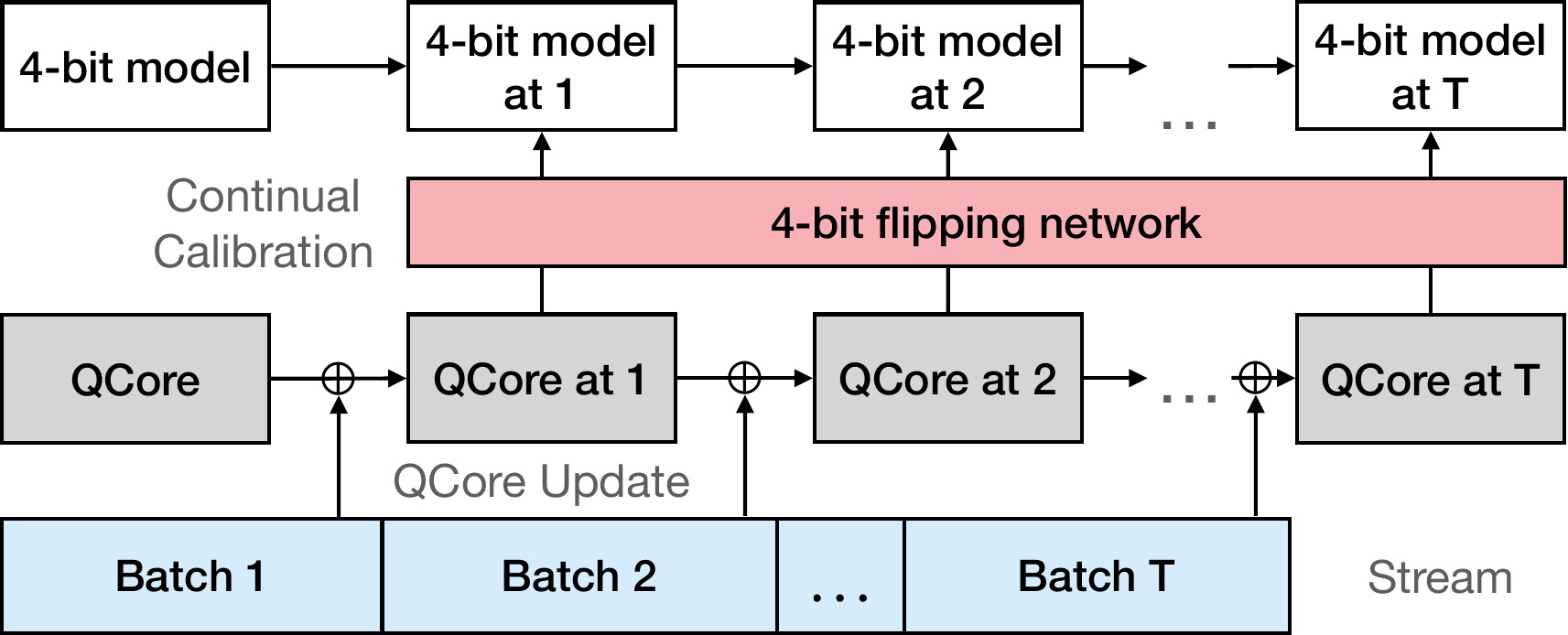}
    \caption{\texttt{QCore} Update, 4-bit Model, $T$ Stream Batches. When a batch from the data stream arrives, the \texttt{QCore} is updated and the quantized model is calibrated by the bit-flipping network accordingly. }
    \label{fig:stream_coreset}
\end{figure}

Once models are deployed on different edge devices, they are exposed to different environments. Each quantized model uses its own incoming streaming data for calibration using the bit-flipping network, as exemplified at the bottom of Figure~\ref{fig:stream_coreset}, for a 4-bit model and a stream.
However, if only new data is taken into account, the model may loose its previous knowledge, a condition called catastrophic forgetting~\cite{LangeAMPJLST22}.

Thus, as \texttt{QCore} is already available on the edge, we use it to prevent catastrophic forgetting. Furthermore, we adjust the \texttt{QCore} to incorporate knowledge from new batches, allowing it to capture both the previous and new domains. Thus, each batch of incoming data is combined with the previous \texttt{QCore} to obtain an updated \texttt{QCore} that is then used for updating the model. Also, since streams will vary across deployments, the \texttt{QCore} will be specialized for each stream and deployment. For instance, the \texttt{QCore} depicted in Figure~\ref{fig:stream_coreset} will be customized for that specific stream.
This approach can be compared to classical continual learning~\cite{LangeAMPJLST22,abs-2302-00487}, where a buffer is used to store knowledge from previous batches. However, a key difference between a buffer and \texttt{QCore} is that \texttt{QCore} integrates the original data and the buffer in a single data structure.

Updates to \texttt{QCore} follows an approach similar to how \texttt{QCores} are built initially, using a distribution of quantization misses. However, when updating \texttt{QCore}, the process is specific to each model, so each stream only adjusts its own \texttt{QCore}. The complete process is detailed in Algorithm~\ref{alg:update_coreset}.
As in Algorithm~\ref{alg:generate_coreset}, the change in the label for each example is evaluated in relation to its correct label across the inference iterations, as can be seen in lines~\ref{line:inference_init}--\ref{line:inference_finish}.

\algblock{Input}{EndInput}
\algnotext{EndInput}
\algblock{Output}{EndOutput}
\algnotext{EndOutput}

\begin{algorithm}
\caption{Update \texttt{QCore}.}
\label{alg:update_coreset}
\begin{algorithmic}[1]

\Input : \texttt{QCore} ($\mathcal{D}_c$), Stream batch ($\mathcal{D}_t$), Quantized model ($\mathit{Q}$)
\EndInput
\Output : Updated \texttt{QCore} ($\mathcal{D}_c$)
\EndOutput
\State $\mathit{QuantMisses}[\mathcal{D}_c \cup \mathcal{D}_t] \gets  \emptyset$ 
\State $\mathcal{D}_c^\prime \gets  \mathcal{D}_c \times 
\frac{|\mathcal{D}_t|}{|\mathcal{D}_c|}$ \Comment{Scaled up to $\mathcal{D}_t$ size}
\Statex \tiny{\text{ }}
\For {$s \gets 1,\ldots,$ $\mathit{E}$}
    \For {$x_i \gets x_1,\ldots, x_m \in \mathcal{D}_c^\prime \cup \mathcal{D}_t$}
        \State $\hat{y}_{i} \gets Q(x_i) $ \Comment{Inference during model calibration} \label{line:inference_init} 
            \If {${\mathit{TP}}_{i}^s$ changes from 1 to 0} \Comment{See Eq.~\ref{eqn:indicator}}
                \State $\mathit{QuantMisses}[x_i] \gets \mathit{QuantMisses}[x_i] + 1$ \label{line:inference_finish}
            \EndIf
    \EndFor
\EndFor
\State \Comment{Count the $N_k$ examples with $k$ quantization misses \qquad \qquad}
\State $\{(k,N_k)\} \gets \mathit{Distribution}(\mathit{QuantMisses})$ \label{line:inference_distribution}
\State $\mathcal{D}_c \gets \mathit{Sample}(Size(\mathcal{D}_c), \mathcal{D}_c^\prime \cup \mathcal{D}_t, \{(k,N_k)\})$ \label{line:inference_sampling}
\end{algorithmic}
\end{algorithm}

\texttt{QCore} updates occur in parallel with model calibration, which is covered in the Section~\ref{ssec:flipping}. Therefore, the epochs in Algorithm~\ref{alg:bit_flip_train} reflect the calibration process when, 
with the incoming examples and \texttt{QCore}, the distribution of quantization misses is recalculated to obtain an updated \texttt{QCore} of the same size. 

\noindent
\textit{Complexity:} Algorithm~\ref{alg:update_coreset} counts the number of quantization misses for each epoch $E$ over the examples within the streaming batch and \texttt{QCore}, which are at most $N$. Therefore, its asymptotic complexity is $\mathcal{O}(E \times N)$.

\section{Experiments} \label{sec:experiments}

\subsection{Experimental Setup}

\subsubsection{Data Sets}

We assess the framework by evaluating the proposed method using two different time-series data sets, \textit{USC}~\cite{ZhangS12a} and \textit{DSA}~\cite{BarshanY14}, and one data set of images, \textit{Caltech10}~\cite{GongSSG12}. The time-series data sets consist of sensor readings from human activities. The labels indicate different conditions such as walking, running, cycling, and rowing. 
The image data set consists of images of office equipment such as computers, headphones, keyboards, and phones.

Table~\ref{table:data_sets} presents the details of the utilized data sets. The information included in the table encompasses the number of classes, the training partition of the data sets, the number of domains and the input size for each data set.

\begin{table}[ht!]
    \small
    \centering
    \caption{Data Sets.}
    \vspace*{-1em} 
    \label{table:data_sets}
    \begin{tabular}{ |l|c|c|c|c| } 
    \hline
    \textbf{Data Set} &  \textit{Classes} &  \textit{Train/Val/Test} &  \textit{Domains} & \textit{Input Size} \\ 
    \hhline{|-|-|-|-|-|}
    \textit{DSA} & 19 & 7296/456/1368 & 8 & $125 \times 45\text{ dim}$ \\
    \textit{USC} & 12 & 4277/269/807 & 14 & $500 \times 6\text{ dim}$ \\
    \textit{Caltech10} & 10 & 2026/126/381 & 4 & $256 \times 256 \times 3\text{ filter}$ \\
    \hline
    \end{tabular}
\end{table}

Each data set can be grouped with another data set as source or target domains, allowing us to create (source, target) pairs from all possible combinations of domains. In this way, we can simulate a continual learning setting where domain shifts happen. For model training and initial calibration, we use the source domain, while the target domain, which may have a different data distribution, is used for testing the continual calibration. This setup resembles situations where concept drift occurs. 
For example, in the \textit{Caltech10} data set, the four domains are \textit{Amazon}, \textit{Caltech}, \textit{DSLR}, and \textit{Webcam}. Therefore, we can train a model using \textit{DSLR} and test the calibration using \textit{Amazon}, indicated with an arrow as \textit{DSLR} $\rightarrow$ \textit{Amazon}. In the case of time-series, we use the number of subjects to indicate the change in domain, e.g., \textit{Subj. 1} $\rightarrow$ \textit{Subj. 2}.

The continual learning setting is built with the target domain divided into 10 stream batches. These batches are fed into the model sequentially as a stream. Upon receiving each batch, \texttt{QCore} is updated and the model is calibrated using the bit-flipping network. After calibration, the model is evaluated on the corresponding test set for each batch, each one representing one-tenth of the testing set in the target domain.

\subsubsection{Metrics}

To evaluate performance, we consider \textit{Accuracy}, which is the proportion of testing examples where the class with the highest probability matches the correct label. For an overall evaluation of the continual learning, we use the average \textit{Accuracy} across all batches. To compare the computational requirements of the model, we consider \textit{Running Time} of each calibration and the size each evaluated data structure, such as \texttt{QCore}, as a proxy for the memory consumption.

\subsubsection{Baselines}
To evaluate the construction of \texttt{QCore}, we consider models at various levels of quantization and compare them to a random subset. We are unable to assess other strategies for building other types of subsets as they do not support quantized models in a continual learning setting, as explained in Section~\ref{sec:related_work}.
Then, the overall \texttt{QCore} model is compared with six state-of-the-art continual learning methods. These methods primarily rely on buffer strategies to retain knowledge from previous batches and utilize back-propagation for making model adjustments. To ensure fair comparison, we keep the sizes of \texttt{QCore} and the buffers the same. 
We consider the following baselines:

\begin{itemize}
    \item Average Gradient Episodic Memory (\texttt{A-GEM}) \cite{ChaudhryRRE19}: This is a rehearsal method that employs a small buffer to sort the gradients of the model after each batch.
    \item Dark Experience Replay (\texttt{DER}) \cite{BuzzegaBPAC20}: This method employs knowledge distillation~\cite{HintonVD15} for rehearsal learning, keeping track of and matching the outputs of each batch.
    \item Dark Experience Replay++ (\texttt{DER++}) \cite{BuzzegaBPAC20,BoschiniBBPC23}: This method introduces a buffer into the \texttt{DER} method to prevent sudden shifts during training.
    \item Experience Replay (\texttt{ER}) \cite{RiemerCALRTT19}: This is the original rehearsal method that maintains a buffer with old samples that are used together with new examples to train the model.
    \item Experience Replay with Asymmetric Cross\Hyphdash Entropy (\texttt{ER\Hyphdash ACE}) \cite{CacciaAATPB22}: This method introduces a training rule into the \texttt{ER} methods that enforces the change of new examples to the previous learning.
    \item Efficient Data Management for Stream Learning (\texttt{Camel}) \cite{LiSC22}: This method introduces a training subset for compressing the incoming data while it keeps a buffer to prevent forgetting previous knowledge.
    \item Deep Compression (\texttt{DeepC}) \cite{HanMD15}: This method employs a three-stage compression encompassing pruning, quantization, and Huffman encoding.
\end{itemize}

\subsubsection{Implementation Details}

The proposed method is implemented using Python 3.8.0 and the machine learning architecture PyTorch 1.13.0. The source code is publicly available at \url{https://github.com/decisionintelligence/QCore}. All models are tested under Ubuntu 22.04.2 using Titan RTX GPUs with 24GB VRAM and an Intel Xeon W-2155 with 128GB RAM.

For all methods, a validation set is used to adjust the hyper-parameters, following the common practice in training machine learning methods. The framework handles different type of models and tested 
using the classification models \texttt{InceptionTime}~\cite{FawazLFPSWWIMP20} and \texttt{OmniScaleCNN}~\cite{TangLLZB022} for time-series and \texttt{ResNet18}~\cite{HeZRS16} and \texttt{VGG16}~\cite{SimonyanZ14a} for image data.
Reported results are average results across five runs using different random seeds to ensure a fair evaluation. 
When using back-propagation, the models in the streaming setting are trained for 200 epochs with a learning rate of 0.01, using the Stochastic Gradient Descent optimizer, and with a batch size of 64. The \texttt{QCore} size, or the corresponding buffer, is selected at 30 examples, in order to keep it small while it approximates the size of an evenly distributed subset where 2-3 examples per class are included.

\subsection{Experimental Results}

\subsubsection{Quantization-Aware QCore}

\begin{figure}[ht]
\small
\centering
\begin{subfigure}{\linewidth}\
\centering
\begin{tikzpicture}
        \matrix [matrix of nodes,nodes={anchor=north},fill=white,draw,inner sep=1pt,row sep=1pt,
        xshift=0pt,yshift=0pt,font=\footnotesize] {
        \ref{forget-4} Core 2
        \ref{forget-8} Core 4 
        \ref{forget-16} Core 8 
        \ref{forget-full} Core 32 (full-precision) \\
        };
        \end{tikzpicture}
\end{subfigure}
\begin{subfigure}{0.5\linewidth}
\vspace*{0.5em}
    \begin{tikzpicture}
        \begin{axis}[
            xlabel=Quantization misses,
            ylabel=Examples,
            xmin=1,
            xmax=9,
            width=1*\linewidth,
            height=0.48*\axisdefaultheight]
            \addplot[black,mark=*,dashed, fill opacity=0.5] table[x=Forgetting, y=Full] {Figures/Data/Activities_1.txt}; \label{forget-full-eog} 
            \addplot[blue, mark=triangle, fill opacity=0.2] table[x=Forgetting, y=2] {Figures/Data/Activities_1.txt}; \label{forget-4-eog} 
            \addplot[red, mark=*, fill opacity=0.2] table[x=Forgetting, y=4] {Figures/Data/Activities_1.txt}; \label{forget-8-eog} 
            \addplot[black, mark=square, fill opacity=0.2] table[x=Forgetting, y=8] {Figures/Data/Activities_1.txt}; \label{forget-16-eog}   
        \end{axis}
    \end{tikzpicture}
    \caption{\textit{DSA} Subj. 1.}
    \label{subfig:forgettingEOGH}
\end{subfigure}
\hspace*{-1.5ex}
\begin{subfigure}{0.5\linewidth}
    \begin{tikzpicture}
        \begin{axis}[
            xlabel=Quantization misses,
            ylabel=Examples,
            xmin=1,
            xmax=9,
            width=1*\linewidth,
            height=0.48*\axisdefaultheight]
            \addplot[black,mark=*,dashed, fill opacity=0.5] table[x=Forgetting, y=Full] {Figures/Data/USC_1.txt}; \label{forget-full} 
            \addplot[blue, mark=triangle, fill opacity=0.2] table[x=Forgetting, y=2] {Figures/Data/USC_1.txt}; \label{forget-4} 
            \addplot[red, mark=*, fill opacity=0.2] table[x=Forgetting, y=4] {Figures/Data/USC_1.txt}; \label{forget-8} 
            \addplot[black, mark=square, fill opacity=0.2] table[x=Forgetting, y=8] {Figures/Data/USC_1.txt}; \label{forget-16}   
        \end{axis}
    \end{tikzpicture}
    \caption{\textit{USC} Subj. 6.}
    \label{subfig:forgettingECG1}
\end{subfigure}
\vspace*{-0.5em}
\caption{Quantization Miss Distributions by Bits.}
\label{fig:forgetting_distribution}
\end{figure}
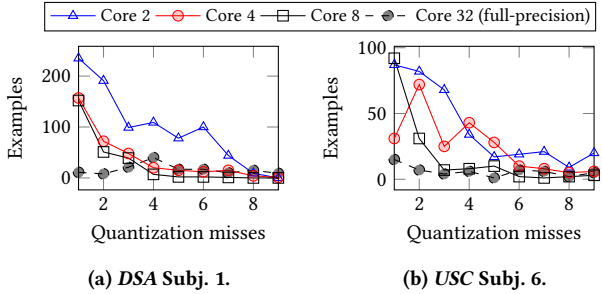

To understand the differences between the possible \texttt{QCores} at various levels of quantization, we calculated the distribution of quantization misses for three different bit-width configurations. We also compute the misses for the full-precision model, which are solely attributed to training, not quantization. For instance, when training a 4-bit model, the quantization misses for the training set are shown in the distribution labeled \textit{Core 4}. This is illustrated in Figures~\ref{subfig:forgettingEOGH} and \ref{subfig:forgettingECG1} for a subject in two data sets. We have found similar observations for other data sets and subjects. 

The comparison of the distributions in both data sets reveals a considerable difference between the full-precision model and the quantized models. This indicates that the difficulty level of examples is affected by quantization. In the full-precision case, the total number of quantization misses is relatively low, which 
suggests that it may not be a reliable indicator when training quantized models. This is because it does not include enough examples where quantized models have more difficulty in processing, such as boundary cases, and therefore, have reduced margin to calibrate properly.

Even when there are similarities between distributions, such as for Core 4 and 8 for the \textit{DSA Subj. 1}, the number of quantization misses is consistently higher for Core 4. This difference becomes more acute as the level of quantization increases, as shown by the results for Core 2. This highlights the importance of considering multiple quantization levels when constructing \texttt{QCore}, as this helps identify the difficulty of examples across different models. Consequently, the \texttt{QCores} can be used to refine different models by including consistently difficult examples, rather than focusing on outliers specific to a particular level of quantization.

Using the quantization miss distributions for two subjects in \textit{DSA}, we constructed different \texttt{QCores} to assess their effectiveness in calibrating quantized models. We examined three quantization levels (2, 4, and 8 bits) with three types of \texttt{QCores} of size 30 for each level, as outlined in Table~\ref{table:coreset_types}. The first type, \textit{Core j}, computes the examples using the same quantization level $j$ as the model, utilizing only one of the quantized distributions, shown as $\mathit{Weight}_j$ in Algorithm~\ref{alg:generate_coreset} and exemplified in Figures~\ref{subfig:forgettingEOGH} and \ref{subfig:forgettingECG1}. The second type, \textit{Core 32}, employs the full-precision network to compute the \texttt{QCore}, employing only the \textit{Core 32} distribution from the figures. Then, \texttt{QCore} aggregates the distributions for all three quantization levels, as described in Algorithm~\ref{alg:generate_coreset}, ensuring the identification of consistently difficult examples across multiple models. For reference, we also include a \textit{Random} subset of the same size.

\begin{table}[ht!]
\addtolength{\tabcolsep}{-1.5pt}
    \small
    \centering
    \caption{Average Accuracy of Quantized Models by Subset Type. \textit{DSA}. Subset Size 30.}
    \label{table:coreset_types}
    \begin{tabular}{ |l|*{8}{c|} } 
    \hline
 \textbf{Subset} & \textit{2-bit} & \textit{4-bit} & \textit{8-bit} & \textit{Avg.} & \textit{2-bit} & \textit{4-bit} & \textit{8-bit} & \textit{Avg.}\\ 
    \hline
 & \multicolumn{4}{l|}{\textit{Subj. 1} $\rightarrow$ \textit{Subj. 2}} & \multicolumn{4}{l|}{\textit{Subj. 1} $\rightarrow$ \textit{Subj. 3}} \\
\hline
	
\texttt{Core 2} & \textbf{0.606} & 0.440 & 0.510 & 0.519 & \textbf{0.538} & 0.385 & 0.418 & 0.447 \\ 
\texttt{Core 4} & 0.374 & \textbf{0.713} & 0.524 & 0.537 & 0.319 & \textbf{0.637} & 0.440 & 0.465 \\ 
\texttt{Core 8} & 0.418 & 0.538 & \textbf{0.719} & 0.558 & 0.363 & 0.407 & \textbf{0.717} & 0.495 \\ 
\texttt{Core 32} & 0.414 & 0.510 & 0.584 & 0.503 & 0.448 & 0.467 & 0.562 & 0.493 \\ 
\texttt{Random} & 0.414 & 0.524 & 0.538 & 0.492 & 0.453 & 0.480 & 0.524 & 0.486 \\ 
\texttt{QCore} & 0.604 & 0.709 & 0.714 & \textbf{0.676} & 0.516 & 0.632 & 0.703 & \textbf{0.617} \\ 
   \hline
    \end{tabular}
\end{table}

\begin{table*}[ht!]
\addtolength{\tabcolsep}{-1.2pt}
    \small
    \centering
    \caption{Average Accuracy of Quantized Models in a Continual Learning Setting. \textit{DSA} and \textit{USC}, \texttt{QCore}/Buffer Size 30.}
    \label{table:baselines_new}
    \begin{tabular}{ |c|l|*{18}{c|} } 
    \hline
 &  \multirow{2}*{\textbf{Model}}  & \multicolumn{9}{c|}{\textit{InceptionTime}} & \multicolumn{9}{c|}{\textit{OmniScaleCNN}} \\
    \cline{3-20}
& & \textit{2-bit} & \textit{4-bit} & \textit{8-bit} & \textit{2-bit} & \textit{4-bit} & \textit{8-bit} & \textit{2-bit} & \textit{4-bit} & \textit{8-bit} & \textit{2-bit} & \textit{4-bit} & \textit{8-bit} & \textit{2-bit} & \textit{4-bit} & \textit{8-bit} & \textit{2-bit} & \textit{4-bit} & \textit{8-bit} \\ 
    \hline
& & \multicolumn{3}{l|}{\textit{Subj. 1} $\rightarrow$ \textit{Subj. 2}} & \multicolumn{3}{l|}{\textit{Subj. 1} $\rightarrow$ \textit{Subj. 3}} & \multicolumn{3}{l|}{\textit{Overall Average}} & \multicolumn{3}{l|}{\textit{Subj. 4} $\rightarrow$ \textit{Subj. 5}} & \multicolumn{3}{l|}{\textit{Subj. 4} $\rightarrow$ \textit{Subj. 6}} & \multicolumn{3}{l|}{\textit{Overall Average}}\\
     \hline
\parbox[t]{2mm}{\multirow{8}{*}{\rotatebox[origin=c]{90}{\textit{DSA}}}} 
& \texttt{A-GEM} & 0.232 & 0.527 & 0.557 & 0.194 & 0.552 & 0.565 & 0.281 & 0.429 & 0.456 & 0.199 & 0.546 & 0.579 & 0.263 & 0.550 & 0.573 & 0.371 & 0.488 & 0.499 \\ 
& \texttt{DER} & 0.509 & 0.520 & 0.557 & 0.446 & 0.653 & 0.657 & 0.441 & 0.506 & 0.530 & 0.463 & 0.524 & 0.525 & 0.546 & \textbf{0.651} & 0.656 & 0.506 & 0.549 & 0.559 \\ 
& \texttt{DER++} & 0.496 & 0.537 & 0.554 & 0.425 & \textbf{0.658} & 0.666 & 0.432 & 0.497 & 0.522 & 0.505 & 0.573 & 0.579 & 0.586 & 0.639 & 0.646 & 0.519 & 0.555 & 0.563 \\ 
& \texttt{ER} & 0.502 & 0.521 & 0.553 & 0.433 & 0.657 & 0.668 & 0.445 & 0.511 & 0.535 & 0.499 & 0.538 & 0.539 & 0.568 & 0.649 & 0.673 & 0.508 & 0.562 & 0.576 \\ 
& \texttt{ER-ACE} & 0.471 & 0.527 & 0.545 & 0.401 & 0.649 & 0.662 & 0.446 & 0.515 & 0.537 & 0.490 & 0.521 & 0.532 & 0.554 & 0.620 & 0.642 & 0.503 & 0.543 & 0.551 \\ 
& \texttt{Camel} & 0.546 & 0.652 & 0.662 & 0.510 & 0.541 & 0.592 & 0.492 & 0.535 & 0.558 & 0.210 & 0.288 & 0.341 & 0.177 & 0.198 & 0.243 & 0.501 & 0.527 & 0.596 \\ 
& \texttt{DeepC} & 0.455 & 0.482 & 0.535 & 0.373 & 0.387 & 0.405 & 0.479 & 0.525 & 0.556 & 0.160 & 0.204 & 0.233 & 0.269 & 0.281 & 0.358 & 0.301 & 0.335 & 0.352 \\ 
& \texttt{QCore} & \textbf{0.604} & \textbf{0.709} & \textbf{0.714} & \textbf{0.516} & 0.632 & \textbf{0.703} & \textbf{0.530} & \textbf{0.581} & \textbf{0.609} & \textbf{0.507} & \textbf{0.580} & \textbf{0.598} & \textbf{0.604} & 0.648 & \textbf{0.675} & \textbf{0.576} & \textbf{0.606} & \textbf{0.717} \\ 
    \hline
 & & \multicolumn{3}{l|}{\textit{Subj. 6} $\rightarrow$ \textit{Subj. 7}} & \multicolumn{3}{l|}{\textit{Subj. 6} $\rightarrow$ \textit{Subj. 8}} & \multicolumn{3}{l|}{\textit{Overall Average}} & \multicolumn{3}{l|}{\textit{Subj. 10} $\rightarrow$ \textit{Subj. 11}} & \multicolumn{3}{l|}{\textit{Subj. 10} $\rightarrow$ \textit{Subj. 12}} & \multicolumn{3}{l|}{\textit{Overall Average}}\\
     \hline
\parbox[t]{2mm}{\multirow{8}{*}{\rotatebox[origin=c]{90}{\textit{USC}}}} 
& \texttt{A-GEM} & 0.222 & 0.497 & 0.614 & 0.054 & 0.464 & 0.534 & 0.131 & 0.369 & 0.459 & 0.268 & 0.443 & 0.673 & 0.154 & 0.407 & 0.647 & 0.173 & 0.397 & 0.456 \\ 
& \texttt{DER} & 0.363 & 0.489 & 0.716 & 0.155 & 0.350 & 0.445 & 0.235 & 0.421 & 0.489 & 0.230 & 0.561 & 0.727 & 0.142 & 0.573 & 0.650 & 0.251 & 0.440 & 0.495 \\ 
& \texttt{DER++} & 0.358 & 0.487 & 0.737 & 0.155 & 0.377 & 0.458 & 0.239 & 0.420 & 0.489 & 0.193 & 0.545 & 0.711 & 0.161 & 0.591 & 0.626 & 0.257 & 0.448 & 0.492 \\ 
& \texttt{ER} & 0.363 & 0.550 & 0.716 & 0.152 & 0.360 & 0.448 & 0.234 & 0.424 & 0.490 & 0.202 & \textbf{0.586} & 0.736 & 0.139 & 0.602 & 0.670 & 0.248 & 0.447 & 0.497 \\ 
& \texttt{ER-ACE} & 0.358 & 0.539 & 0.700 & 0.161 & 0.377 & 0.465 & 0.242 & 0.423 & 0.493 & 0.193 & 0.543 & 0.730 & 0.150 & 0.595 & 0.628 & 0.261 & 0.452 & 0.500 \\ 
& \texttt{Camel} & 0.646 & 0.657 & 0.732 & 0.383 & 0.424 & 0.452 & 0.333 & 0.377 & 0.454 & 0.243 & 0.244 & 0.306 & 0.198 & 0.218 & 0.223 & 0.483 & 0.496 & 0.529 \\ 
& \texttt{DeepC} & 0.616 & 0.652 & 0.704 & 0.435 & 0.491 & 0.595 & 0.336 & 0.407 & 0.475 & 0.137 & 0.181 & 0.213 & 0.061 & 0.080 & 0.157 & 0.184 & 0.194 & 0.224 \\ 
& \texttt{QCore} & \textbf{0.783} & \textbf{0.846} & \textbf{0.870} & \textbf{0.609} & \textbf{0.696} & \textbf{0.696} & \textbf{0.463} & \textbf{0.524} & \textbf{0.621} & \textbf{0.448} & \textbf{0.586} & \textbf{0.737} & \textbf{0.448} & \textbf{0.609} & \textbf{0.710} & \textbf{0.501} & \textbf{0.516} & \textbf{0.556} \\  
   \hline
    \end{tabular}
\end{table*}

The results in Table~\ref{table:coreset_types} offer two important insights: first, the construction of \texttt{QCore} based on the quantization miss distributions is a good proxy for adjusting quantized models, and it performs consistently well for different bit-widths and achieve the best average accuracy; and second, the original distribution is reproduced closely, even at reduced \texttt{QCore} sizes, such as 30 examples. The results show that the proposed strategy enables the creation of highly compressed subsets that can be deployed on edge devices. Additionally, the \textit{Random} subset performs the worst in several cases, close to the non-specific \textit{Core 32}, indicating that it is not a good strategy for defining a quantization-aware subset.

The results indicate that a subset computed with the same bit-width as the quantized model performs better than the alternatives; for example, for the 4-bit quantized model, \textit{Core 4} performs better than \textit{Random} and the non-specific subsets. This outcome is expected, as the subset is each designed specifically for a particular model. However, usability and scalability are limited as these subsets can only be used for a specific model; for example, \textit{Core 4} is too specific for use on 8-bit models. This is evident from its lower results and applies to any case where Core $j$ does not correspond to the $k$-bit model, i.e., $k\neq j$. Addressing this issue, \texttt{QCore} offers comparatively good performance and can be applied in all quantized settings, as the average results shows. This means that only one \texttt{QCore} is needed to calibrate models with different bit-widths, which is desirable when deploying models on edge devices with varying constraints. Additionally, the accuracy achieved using \texttt{QCore} is higher than that achieved when using the full-precision \textit{Core 32}. This underscores the importance of having a \texttt{QCore} that is quantization-aware and can identify challenging examples for multiple quantized models.

\subsubsection{Continual Calibration Evaluation}

When comparing our proposal with other continual learning methods, we consider the average accuracy for ten batches across three quantization levels: 2, 4, and 8 bits. As mentioned, the models are trained in one domain and calibrated in another, indicated by arrows. 
We conducted experiments for all possible combinations of domains, totaling 56 scenarios for \textit{DSA}, 182 for \textit{USC}, and 12 for \textit{Caltech10}. Due to space limitation, we only show an excerpt of the results. 
This excerpt consists of four randomly chosen training domains coupled with their next two calibration domains---see Table~\ref{table:baselines_new}, and the overall average results. For \textit{Caltech10}, the average results are shown in Table~\ref{table:caltech_avg}.
The best results are highlighted in bold.

When compared with the continual learning baselines for time-series data in Table~\ref{table:baselines_new}, \texttt{QCore} achieves the best results by a significant margin. For all methods, the accuracy increases as the bit-width increases, which is expected because the models have fewer constraints on their parameters. Then, \textit{DSA} has two cases: \textit{Subj. 1} $\rightarrow$ \textit{Subj. 3} and \textit{Subj. 4} $\rightarrow$ \textit{Subj. 6} both under 4-bit, where the performance of \texttt{QCore} is not the best. When examining these scenarios in detail, it seems that in both cases, one particular batch affect significantly the performance of \texttt{QCore}, thereby decreasing the overall average result. This condition is infrequent, as shown by all the other results.
In addition, the performance of the baselines on both \textit{DSA} and \textit{USC} is relatively similar, except for \texttt{A-GEM} that consistently performs the worst in almost all cases. 

For the \textit{Caltech10} image data set, Table~\ref{table:caltech_avg} demonstrates a similar outcome to the results on time-series data. The performance of \texttt{QCore} outperforms the baselines in every evaluated scenario. This evaluation also highlights the applicability of the framework in a relatively uncommon streaming setting with images, supporting its usability across substantially different settings.

\begin{table}[ht!]
\addtolength{\tabcolsep}{-1pt}
    \small
    \centering
    \caption{Average Accuracy of Quantized Models in a Continual Learning Setting. \textit{Caltech10}, \texttt{QCore}/Buffer Size 30.}
    \vspace*{-1em}
    \label{table:caltech_avg}
    \begin{tabular}{ |l|*{6}{c|} } 
    \hline
    \multirow{2}*{\textbf{Model}}  & \multicolumn{3}{c|}{\textit{ResNet18}} & \multicolumn{3}{c|}{\textit{VGG16}} \\
    \cline{2-7}
    & \textit{2-bit} & \textit{4-bit} & \textit{8-bit} & \textit{2-bit} & \textit{4-bit} & \textit{8-bit} \\ 
    \hline
\texttt{A-GEM} & 0.329 & 0.355 & 0.364 & 0.079 & 0.096 & 0.114 \\ 
\texttt{DER} & 0.345 & 0.363 & 0.368 & 0.117 & 0.126 & 0.132 \\ 
\texttt{DER++} & 0.341 & 0.356 & 0.358 & 0.108 & 0.119 & 0.135 \\ 
\texttt{ER} & 0.353 & 0.367 & 0.369 & 0.122 & 0.140 & 0.174 \\ 
\texttt{ER-ACE} & 0.346 & 0.360 & 0.362 & 0.108 & 0.126 & 0.140 \\ 
\texttt{Camel} & 0.348 & 0.363 & 0.369 & 0.148 & 0.162 & 0.174 \\ 
\texttt{DeepC} & 0.346 & 0.361 & 0.367 & 0.148 & 0.160 & 0.182 \\ 
\texttt{CoreQ} & \textbf{0.399} & \textbf{0.414} & \textbf{0.431} & \textbf{0.181} & \textbf{0.187} & \textbf{0.202} \\ 
\hline
    \end{tabular}
\end{table}

\subsubsection{Ablation Study} \label{sssec:ablation}

To assess the importance of the components of \texttt{QCore}, we examine the effects of removing them during model calibration in a continual learning setting with multiple batches.
We consider two scenarios: removing the \texttt{QCore} update (Algorithm~\ref{alg:update_coreset} in Section~\ref{ssec:coreset_update}) component (\texttt{NoUpda}) and removing the bit-flipping (Section~\ref{ssec:flipping}) component (\texttt{NoBF}), comparing them to the complete method (\texttt{QCore}).

The evaluation considers ten batches, to show how the method evolves when processing the complete streaming domain set. After processing these batches, their average is computed. Table~\ref{table:ablation} reports the accuracy for the two pairs of domains shown in Table~\ref{table:baselines_new} for \texttt{InceptionTime} in both data sets using a quantized model of 4 bits. Similar results are observed for other bit-widths and settings.

The results show that, on average, \texttt{QCore} achieves the highest accuracy. 
This implies that the method can quickly adjust when a new batch is introduced, thanks to its bit-flipping mechanism. Additionally, it effectively retains past knowledge to prevent catastrophic forgetting, as evidenced by the results. 
When considering the overall execution time, the relatively small differences highlight the efficiency of \texttt{QCore}, due the low overhead of its components.

\begin{table}[ht!]
    \small
    \centering
    \caption{Ablation Study of Quantized Models by Incoming Batches. Accuracy, 4-bit, Subset Size 30.}
    \label{table:ablation}
    \begin{tabular}{ |l|*{6}{c|} } 
    \hline
  & \texttt{NoUpda} & \texttt{NoBF} & \texttt{QCore} & \texttt{NoUpda} & \texttt{NoBF} & \texttt{QCore} \\ 
\hline
\textit{Batch} & \multicolumn{3}{l|}{\textit{DSA}: \textit{Subj. 1} $\rightarrow$ \textit{Subj. 2}}  & \multicolumn{3}{l|}{\textit{USC}: \textit{Subj. 6} $\rightarrow$ \textit{Subj. 7}}  \\
\hline
 
\textit{1} & 0.659 & 0.473 & \textbf{0.675} & 0.435 & 0.387 & \textbf{0.957} \\ 
\textit{2} & 0.505 & 0.429 & \textbf{0.582} & 0.304 & 0.419 & \textbf{0.652} \\ 
\textit{3} & 0.692 & 0.538 & \textbf{0.765} & 0.391 & 0.387 & \textbf{0.965} \\ 
\textit{4} & 0.604 & 0.571 & \textbf{0.725} & 0.522 & 0.323 & \textbf{0.826} \\ 
\textit{5} & 0.637 & 0.571 & \textbf{0.648} & 0.435 & 0.387 & \textbf{0.652} \\ 
\textit{6} & 0.516 & 0.538 & \textbf{0.747} & 0.478 & 0.355 & \textbf{0.952} \\ 
\textit{7} & 0.571 & 0.440 & \textbf{0.755} & 0.391 & 0.323 & \textbf{0.957} \\ 
\textit{8} & 0.593 & 0.527 & \textbf{0.780} & 0.609 & 0.290 & \textbf{0.909} \\ 
\textit{9} & 0.571 & 0.484 & \textbf{0.670} & 0.217 & 0.452 & \textbf{0.846} \\ 
\textit{10} & 0.363 & 0.549 & \textbf{0.736} & 0.478 & 0.290 & \textbf{0.696} \\ 
\textit{Avg.} & 0.571 & 0.512 & \textbf{0.708} & 0.426 & 0.361 & \textbf{0.841} \\ 
\hline
\textit{Time (s)} & 5.607 & 5.523 & 5.659 & 4.874 & 4.456 & 5.081 \\  
\hline
    \end{tabular}
    \vspace*{-1em}
\end{table}

\subsubsection{\texttt{QCore} Construction} \label{sssec:construction}
To evaluate the construction of \texttt{QCore}, we compare \texttt{QCore} with other sampling strategies and gradient-based subsets using \texttt{InceptionTime} as a backbone and without continual calibration to isolate the sets.
We evaluate three sampling strategies~\cite{cochran1991}. First, maximum entropy that includes the most dissimilar sample in the subset compared to the ones already selected. 
Second, least confidence that adds the sample with the most uncertain similarity to the subset. 
Third, a sampling that assumes that the quantization misses follow a normal distribution. 
The results, at the top in Table~\ref{table:core_types}, show that \texttt{QCore} outperforms them.

We also evaluate other coreset construction methods. These include a variation of \texttt{k-means}~\cite{reference/ml/2017}, which selects the examples close to the centroids, as well as two gradient-based methods. The first, \texttt{GradMatch}~\cite{KillamsettySRDI21}, selects examples dynamically to match the full-gradient at each training step, and the second, \texttt{CRAIG}~\cite{MirzasoleimanBL20}, aims to find the optimal coreset that minimizes the gradient loss using a set cover approximation. 
The results, at the bottom in Table~\ref{table:core_types}, show that \texttt{QCore} performs the best. 

\begin{table}[ht!]
\addtolength{\tabcolsep}{-1pt}
    \small
    \centering
    \caption{Average Accuracy on Coreset Construction Strategies. Subset Size 30.}
    \vspace*{-1em}
    \label{table:core_types}
    \begin{tabular}{ |l|*{6}{c|} } 
    \hline
 & \textit{2-bit} & \textit{4-bit} & \textit{8-bit} & \textit{2-bit} & \textit{4-bit} & \textit{8-bit}\\ 
    \hline
 \textbf{Strategy} & \multicolumn{3}{l|}{\textit{DSA}} & \multicolumn{3}{l|}{\textit{USC}} \\
\hline
	
\texttt{Maximum Entropy} & 0.578 & 0.613 & 0.635 & 0.292 & 0.330 & 0.414 \\ 
\texttt{Least Confidence} & 0.573 & 0.603 & 0.620 & 0.284 & 0.321 & 0.403 \\ 
\texttt{Normal Distrib.} & 0.594 & 0.599 & 0.635 & 0.291 & 0.335 & 0.419 \\ 
\hline
\texttt{k-means} & 0.590 & 0.602 & 0.640 & 0.285 & 0.323 & 0.406 \\ 
\texttt{GradMatch} & 0.592 & 0.603 & 0.641 & 0.292 & 0.325 & 0.409 \\ 
\texttt{CRAIG} & 0.587 & 0.602 & 0.644 & 0.293 & 0.339 & 0.423 \\ 
\hline
\texttt{QCore} & \textbf{0.597} & \textbf{0.614} & \textbf{0.648} & \textbf{0.307} & \textbf{0.354} & \textbf{0.436} \\ 
\hline
    \end{tabular}
\end{table}

\subsubsection{Running Time}  \label{sssec:running_time}

To evaluate the running time, we executed the models independently, so that no other processes were running on the system that could significantly affect the performance. The average end-to-end runtime is shown in Table~\ref{table:average_time} for the 4-bit model in the continual learning setting; results on other quantization levels show similar execution times since the calibration process is equivalent.
When evaluating all the data sets, \texttt{QCore} consistently outperforms all the baselines, with a speed-up of up to three to five times for every case.

\begin{figure}[ht]
\small
\centering
\begin{subfigure}{0.5\linewidth}
    \begin{tikzpicture}
        \begin{axis}[
    xlabel=Epoch,
    ylabel=Accuracy,
    xmin=0,
    xmax=50,
            width=1*\linewidth,
            height=0.5*\axisdefaultheight,
            legend image post style={xscale=0.4},
            legend style={
                    at={(-0.35,1.47)},
                    anchor=north west,
                    legend columns=4,},font=\footnotesize]
            \addplot[blue, dashed,mark=none] table[x=epoch, y=agem] {Figures/Data/EpochsBaselines.txt};
            \addlegendentry{\texttt{A-GEM}}
            \addplot[red, dashed,mark=none] table[x=epoch, y=der] {Figures/Data/EpochsBaselines.txt};
            \addlegendentry{\texttt{DER}}
            \addplot[black, dashed,mark=none] table[x=epoch, y=derpp] {Figures/Data/EpochsBaselines.txt};
            \addlegendentry{\texttt{DER++}}
            \addplot[purple, mark=none] table[x=epoch, y=er] {Figures/Data/EpochsBaselines.txt};
            \addlegendentry{\texttt{ER}}
            \addplot[blue,dotted,mark=none] table[x=epoch, y=er_ace] {Figures/Data/EpochsBaselines.txt};
            \addlegendentry{\texttt{ER-ACE}}
            \addplot[red, mark=none] table[x=epoch, y=camel] {Figures/Data/EpochsBaselines.txt};
            \addlegendentry{\texttt{Camel}}
            \addplot[black, mark=none] table[x=epoch, y=deepc] {Figures/Data/EpochsBaselines.txt};
            \addlegendentry{\texttt{DeepC}}
            \addplot[blue,mark=none] table[x=epoch, y=qcore] {Figures/Data/EpochsBaselines.txt};
            \addlegendentry{\texttt{QCore}}
        \end{axis}
    \end{tikzpicture}
    \caption{Convergence Analysis.}
    \vspace*{-0.5em}
    \label{subfig:epochs}
\end{subfigure}
\hspace*{-1.5ex}
\begin{subfigure}{0.5\linewidth}
    \begin{tikzpicture}
        \begin{axis}[
            xlabel=Buffer/Subset size,
            ylabel=Accuracy,
            width=1*\linewidth,
            height=0.5*\axisdefaultheight,
            legend style={
                    at={(-0.2,1.49)},
                    anchor=north west,
                    legend columns=4,},font=\footnotesize]
            \addplot[blue, mark=triangle,only marks,mark options={scale=1, fill=blue}] table[x=Memory, y=agem_r] {Figures/Data/Memory.txt}; 
            \addlegendentry{\texttt{A-GEM}}
            \addplot[blue, mark=square,only marks] table[x=Memory, y=der] {Figures/Data/Memory.txt}; \addlegendentry{\texttt{DER}}
            \addplot[red, mark=star,only marks] table[x=Memory, y=derpp] {Figures/Data/Memory.txt}; \addlegendentry{\texttt{DER++}}
            \addplot[blue, mark=*,only marks] table[x=Memory, y=er] {Figures/Data/Memory.txt}; \addlegendentry{\texttt{ER}}
            \addplot[purple, mark=none,only marks] table[x=Memory, y=er_ace] {Figures/Data/Memory.txt}; \addlegendentry{\texttt{ER-ACE}}
            \addplot[red, mark=triangle,only marks] table[x=Memory, y=camel] {Figures/Data/Memory.txt};
            \addlegendentry{\texttt{Camel}}
            \addplot[black, mark=square,only marks] table[x=Memory, y=deepc] {Figures/Data/Memory.txt};
            \addlegendentry{\texttt{DeepC}}
            \addplot[blue, mark=star,only marks] table[x=Memory, y=quant] {Figures/Data/Memory.txt};
            \addlegendentry{\texttt{QCore}}
        \end{axis}
    \end{tikzpicture}
    \caption{Memory Consumption.}
    \vspace*{-0.5em}
    \label{subfig:memory}
\end{subfigure}
\caption{Convergence and Memory Consumption Evaluation. \textit{DSA} \textit{Subj. 1} $\rightarrow$ \textit{Subj. 2}, 4-bit.}
\label{fig:efficiency}
\end{figure}
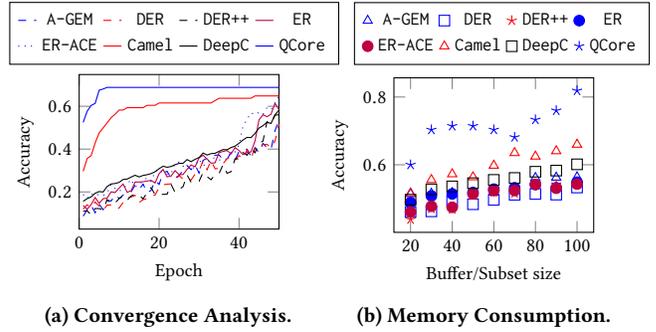

The high efficiency of \texttt{QCore} in terms of execution time is explained by the design of its bit-flipping network.
First, the baselines use back-propagation to update model parameters, which require computing gradients. In contrast, the bit-flipping network only requires
a single inference step to compute an adjustment on model parameters. This improves performance substantially. 
Second, the bit-flipping network uses much less epochs to converge during a calibration compared to baselines using back-propagation, as shown in Figure~\ref{subfig:epochs}. 
It is observed that in less than ten epochs \texttt{QCore} is already stable. That is expected because the bit-flipping network is designed for inference-only purposes. Therefore, the calibration steps are  minimal, 
which is consistent with the convergence analysis in Section~\ref{sssec:flipping_update}. In contrast, all the baselines need to execute gradient computations, which takes more epochs to converge.

\begin{table}[ht!]
\addtolength{\tabcolsep}{-2.5pt}
    \small
    \centering
    \caption{Average End-to-End Running Time per Calibration (seconds), 4-bit, \texttt{QCore}/Buffer Size 30. } 
    \vspace*{-1em}
    \label{table:average_time}
    \begin{tabular}{ |l|*{8}{c|} } 
    \hline
& \texttt{A-GEM} & \texttt{DER} & \texttt{DER++} & \texttt{ER} & \texttt{ER-ACE} & \texttt{Camel} & \texttt{DeepC} & \texttt{QCore} \\ 
\hline
\textit{DSA} & 15.24 & 11.41 & 15.98 & 10.26 & 11.18 & 13.32 & 12.82 & \textbf{3.44} \\ 
\textit{USC} & 13.86 & 10.97 & 15.45 & 11.68 & 10.94 & 14.30 & 12.58 & \textbf{3.43} \\ 
\textit{Calt10} & 113.19 & 100.45 & 137.01 & 117.86 & 101.03 & 159.17 & 113.91 & \textbf{31.83 } \\ 
   \hline
    \end{tabular}
\end{table}

\subsubsection{Memory Consumption}
To compare the memory consumption between \texttt{QCore} and the evaluated baselines, we computed the results for the \textit{Subj. 1} $\rightarrow$ \textit{Subj. 2} \texttt{QCore} with the 4-bit model in the \textit{DSA} data set. We compared these results with those for the baselines across different buffer sizes, as shown in Figure~\ref{subfig:memory}. 
The results show a positive trend, indicating that the models improve as the buffer size increases. 
Even so, most of the baselines show relatively small variations, indicating that they may not be selecting the most appropriate examples given the limited space. \texttt{Camel} performs competitively compared to \texttt{QCore} with closer results for some subset sizes. This illustrates the advantages of keeping a training subset instead of only buffers.
Overall, the comparison highlights the efficiency of \texttt{QCore} in terms of memory consumption, as it is able to identify suitable examples with low memory use.
\section{Related Work} \label{sec:related_work}

We cover the related studies on two relevant aspects: subset building and continual calibration. 

\noindent
\textbf{Subset Building: } The concept of compressing a data set into a representative subset has been the subject of study in the past decade~\cite{FeldmanL11,RosmanVFFR14}, often known as coresets.
We review the relevant studies from two dimensions. First, whether the subset building is quantization-aware, with the purpose of facilitating efficient quantized model calibration. Second, whether the subset can be updated in a stream setting, where the newly arrived data may exhibit different distributions. We summarize the relevant studies into Table~\ref{table:quant_aware}, which shows that QCore is the only study that is quantization-aware and support stream updates. 

\begin{table}[ht!]
\centering
\small
\caption{Related Work on Subset Building.}
\label{table:quant_aware}
\begin{tabular}{cc|c|c|}
\cline{3-4}
&  & \multicolumn{2}{c|}{Stream Update} \\ \cline{3-4}
&  & \cmark &  \xmark \\ \hline
\multicolumn{1}{|c|}{Quantization}  & \cmark  & \texttt{QCore} & - \\ \cline{2-4}
\multicolumn{1}{|c|}{Aware} & \xmark  & \cite{LiSC22,BravermanFLRS23} & ~\cite{CampbellB18,Badoiu08,Hai04,HarPeled11,ChaiL0FM0L023,ManousakasXMC20,HuangSV21,JubranSNF21,Cazenavette00EZ22b,PaparrizosLBHEE21} \\ \hline
\end{tabular}
\end{table}

Coresets mostly focus on compressing data sets for faster training full-precision models, usually relying on geometric closeness~\cite{CampbellB18,Badoiu08,Hai04,HarPeled11,ChaiL0FM0L023} or statistical properties~\cite{ManousakasXMC20}. However, these coreset selection methods may not be suitable for calibrating quantized models, as they only consider a single model, while 
different data samples may have different effects on quantized models with different bit-widths vs. full precision models. In contrast, \texttt{QCore} uses ``quantization misses'' to select a subset that specifically targets at effective calibration of quantized models. 
Next, most existing coreset selection method do not consider how to update coresets when receiving new data, except two studies~\cite{BravermanFLRS23,LiSC22}.  
The empirical shows that \texttt{QCore} outperforms~\cite{LiSC22}, which is original designed for full-precision models. 
We do not compare with \cite{BravermanFLRS23} as it only solves least-mean-squares problems, while we focus on classification. 
There are other techniques for compressing a data set that usually focus on selecting a subset of dimensions ~\cite{IlkhechiCGMFSC20,ElgamalYAMH15,YuAKPZCWI21}. However, these techniques are not applicable in the current setting because the training load will still be large, resulting in the same number of examples. 

\noindent
\textbf{Continual Calibration: } 
In recent years, the idea of continual learning has been studied, with the aim of developing models that can adapt to dynamic environments where stream data keeps arriving~\cite{CacciaAATPB22}. 
We review the relevant studies from two dimensions. Firstly, we examine whether the continual calibration supports quantized models. Secondly, we determine whether the model uses back-propagation that require computing gradients in the calibration, as this can be a costly process when running on edge devices. We have summarized the relevant studies in Table~\ref{table:calibration}.

\begin{table}[ht!]
\addtolength{\tabcolsep}{-1pt}
\centering
\small
\caption{Related Work on Continual Calibration.}
\vspace*{-1em}
\label{table:calibration}
\begin{tabular}{c|c|c|}
\cline{2-3}
& \multicolumn{2}{c|}{Parameters Support} \\ \cline{2-3}
& Full-precision &  Quantized \\ \hline
\multicolumn{1}{|c|}{Calibration with BP} & \cite{LiSC22,RiemerCALRTT19, CacciaAATPB22, BorsosM020, MirzasoleimanBL20,ChaudhryRRE19, ChaudhryGDTL21,BuzzegaBPAC20, BoschiniBBPC23} & \cite{Shi0CF21,LinCLCG020,RenAR21,RavagliaRNCCB21} \\ \hline
\multicolumn{1}{|c|}{Calibration without BP} & \cite{0013FGD22, NiuW0CZZT22} & \texttt{QCore} \\ \hline

\end{tabular}
\end{table}

Most approaches focus on adjusting full-precision parameters using back-propagation ~\cite{RiemerCALRTT19, CacciaAATPB22, BorsosM020, MirzasoleimanBL20,LiSC22}. The primary focus is on enhancing data retention using gradients~\cite{ChaudhryRRE19, ChaudhryGDTL21, MirzasoleimanBL20, KillamsettySRDI21} or knowledge distillation~\cite{BuzzegaBPAC20, BoschiniBBPC23}, but continual calibration is often not supported~\cite{HanMD15}.
Existing continual calibration of quantized models is based on back-propagation, which is expensive on edge devices. In addition, existing studies are often specific for particular micro-controllers~\cite{LinCLCG020,RenAR21,RavagliaRNCCB21} or  particular bit-widths~\cite{RavagliaRNCCB21}, reducing their generality.
Two methods~\cite{0013FGD22, NiuW0CZZT22} exist that do not use BP, but they support only full-precision models, and they suffer catastrophic forgetting. 
\texttt{QCore} enables continual calibrations of quantized models, while preventing the use of expensive BP with the bit-flipping network, making it a perfect on-device edge-ready approach.

\section{Conclusion and Future Work} \label{sec:conclusions}

This paper proposes \texttt{QCore}, a novel and efficient method for on-device training on edge devices with limited resources. \texttt{QCore} employs a quantization-aware subset that compresses the training set and streaming data, identifying the most suitable examples for training a quantized model with a reduced number of bits per model parameter. It also includes a small network that enables continual model calibration without requiring back-propagation, significantly reducing computational costs and enabling implementation on edge devices. The results of the experimental study offer concrete evidence of the effectiveness and efficiency of \texttt{QCore} at classification tasks. The method also demonstrates improvements in running time, while it only requires a small fraction of training and streaming data examples to construct an appropriate subset.

In future work, it is of interest to explore potential guarantees for the \texttt{QCore} and properties of the bit-flipping network. This includes investigating its data independence and the theoretical basis that can validate the empirical findings.

\bibliographystyle{ACM}
\bibliography{Sections/References}

\end{document}